\documentclass{article}

\usepackage{microtype}
\usepackage{graphicx}
\usepackage{subfigure}
\usepackage{booktabs} 
\usepackage[most]{tcolorbox}
\usepackage{fvextra}
\usepackage{booktabs}   
\usepackage{tabularx}
\usepackage{array}
\usepackage{ragged2e}

\DefineVerbatimEnvironment{PromptVerbatim}{Verbatim}{fontsize=\small,breaklines=true,breakanywhere=true}

\newtcolorbox{promptbox}[1]{%
  enhanced,
  breakable,                
  colback=white,
  colframe=black!75,
  boxrule=1.0pt,
  arc=6pt,
  left=10pt,right=10pt,top=8pt,bottom=8pt,
  fonttitle=\bfseries\large,
  title=#1,
  pad at break=4pt,
  segmentation style={solid, black!30},
}

\usepackage{hyperref}
\usepackage{booktabs}
\usepackage{multirow}
\usepackage{xcolor}
\usepackage{colortbl}
\usepackage{enumitem}

\usepackage[accepted]{icml2026}
\usepackage{amsmath}
\usepackage{amssymb}
\usepackage{mathtools}
\usepackage{amsthm}

\usepackage[capitalize,noabbrev]{cleveref}

\theoremstyle{plain}

\theoremstyle{definition}

\theoremstyle{remark}

\usepackage[textsize=tiny]{todonotes}

\definecolor{wdcolor}{RGB}{128, 0, 255}

\definecolor{wdqcolor}{RGB}{255, 0, 0}

\icmltitlerunning{Seeing Clearly without Training}

\begin{document}

\twocolumn[
\icmltitle{Seeing Clearly without Training:\\Mitigating Hallucinations in Multimodal LLMs for Remote Sensing}

\icmlsetsymbol{equal}{$\dagger$}

\begin{icmlauthorlist}
\icmlauthor{Yi Liu}{whu,zgc}
\icmlauthor{Jing Zhang}{whu,zgc,equal}
\icmlauthor{Di Wang}{whu,zgc,equal}
\icmlauthor{Xiaoyu Tian}{cqu}
\icmlauthor{Haonan Guo}{whu,zgc}
\icmlauthor{Bo Du}{whu,zgc,equal}
\end{icmlauthorlist}

\icmlaffiliation{whu}{School of Computer Science, Wuhan University, China}
\icmlaffiliation{zgc}{Zhongguancun Academy, China}
\icmlaffiliation{cqu}{School of Computer Science, Chongqing University, China}

\icmlcorrespondingauthor{Jing Zhang}{jingzhang.cv@gmail.com}
\icmlcorrespondingauthor{Di Wang}{d\_wang@whu.edu.cn}
\icmlcorrespondingauthor{Bo Du}{dubo@whu.edu.cn}

\vskip 0.3in
]

\printAffiliationsAndNotice{$\dagger$ Corresponding author.}

\begin{abstract}
{
Multimodal large language models (MLLMs) suffer from pronounced hallucinations in remote sensing visual question-answering (RS-VQA), primarily caused by visual grounding failures in large-scale scenes or misinterpretation of fine-grained small targets. To systematically analyze these issues, we introduce RSHBench, a protocol-based benchmark for fine-grained diagnosis of factual and logical hallucinations. To mitigate grounding-induced factual hallucinations, we further propose \textbf{R}elative \textbf{A}ttention-\textbf{D}riven \textbf{A}ctively \textbf{R}easoning (RADAR), a training-free inference method that leverages intrinsic attention in MLLMs to guide progressive localization and fine-grained local reasoning at test time. Extensive experiments across diverse MLLMs demonstrate that RADAR consistently improves RS-VQA performance and reduces both factual and logical hallucinations. Code and data will be publicly available at
\href{https://github.com/MiliLab/RADAR}{\texttt{https://github.com/MiliLab/RADAR}}.
}

\end{abstract}

\section{Introduction} 
\begin{figure}[!t]

    \centering

    \includegraphics[width=1\linewidth]{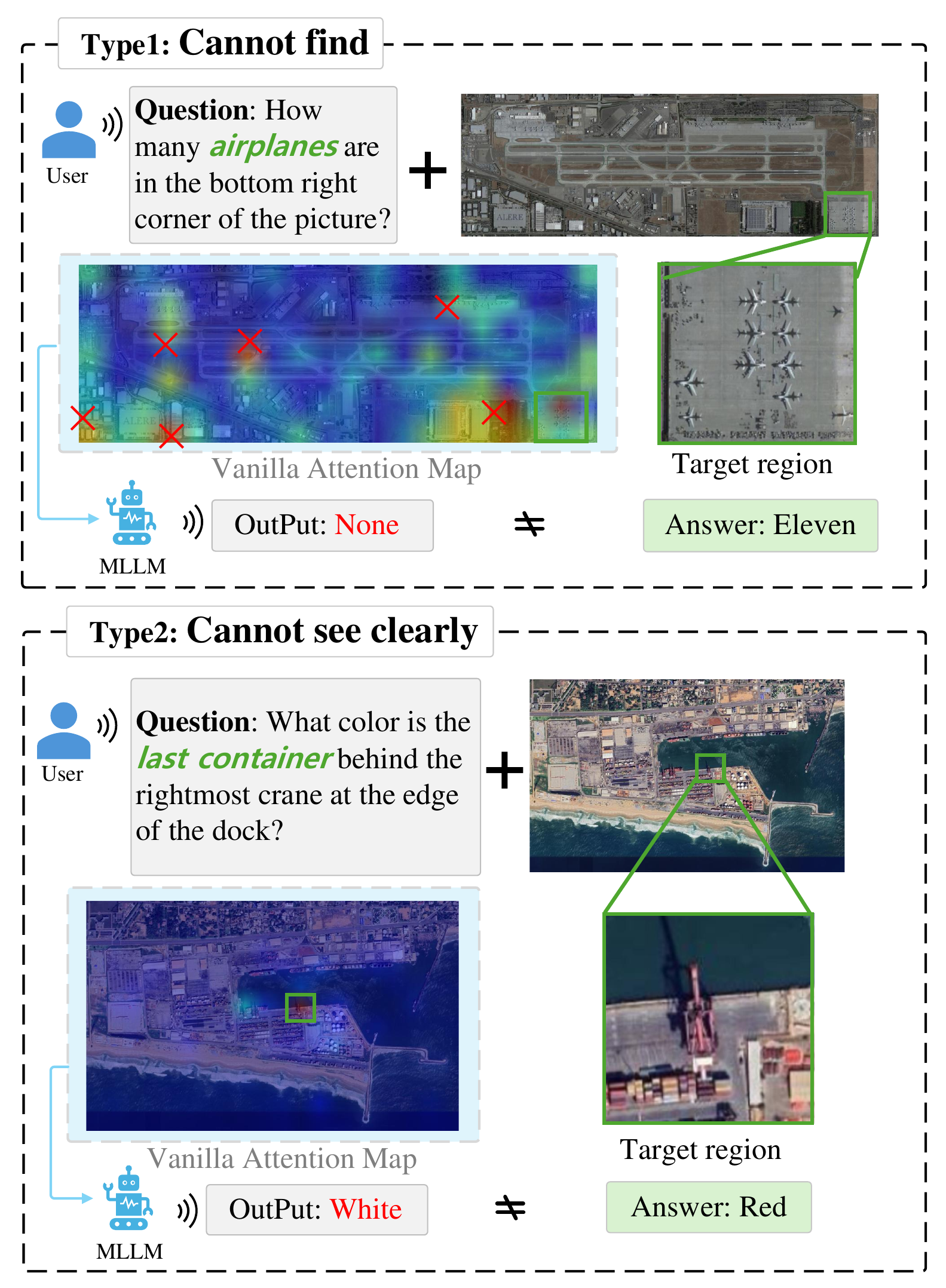} 
\caption{\textbf{Two grounding failures underlying RS-VQA hallucinations.}
{Type~1 (Cannot find):} model attention becomes diffuse and is distracted by irrelevant regions, resulting in missed target localization.
{Type~2 (Cannot see clearly):} attention covers the correct region, but the visual evidence is too small or ambiguous for fine-grained recognition, leading to incorrect predictions.}
\vspace{-0.2cm}

    \label{fig:motivation}

\end{figure}

In recent years, MLLMs~\cite{hurst2024gpt, touvron2023llama, yang2025qwen3, liu2023visual} have demonstrated remarkable enhancements in general visual perception and reasoning capabilities. These developments have concurrently invigorated the fields of Earth science and remote sensing data processing~\cite{11159545,rs_agricultural,rs_urban,rs_damage}. 
{Nevertheless, they continue to exhibit substantial limitations when applied to specialized remote sensing tasks such as RS-VQA.}

This gap is primarily attributed to two fundamental factors. First, since remote sensing imagery typically spans a large spatial extent~\cite{DOTA,7560644}, MLLMs struggle to precisely localize regions that are truly relevant to the query. Second, as scenes are often characterized by substantial background clutter and densely distributed small targets~\cite{Xu_2025_ICCV}, fine-grained recognition becomes particularly challenging. 
{
As a result, MLLMs may rely on language priors in the absence of reliable visual evidence~\cite{halucination_language}, leading to diverse hallucination issues. More critically, such perceptual errors can propagate through the reasoning process, rendering it increasingly fragile~\cite{hallucination_snowballing}.
}

{
As illustrated in Figure~\ref{fig:motivation}, hallucinations in RS-VQA primarily arise from two types of grounding failures.
In Type~1 (Localization Failure), the question hinges on a small region within a large scene, yet the model’s attention becomes diffuse and is distracted by irrelevant regions, causing the true target area to be missed.
In Type~2 (Recognition Failure), the model attends to the correct region, but the visual evidence is too small or ambiguous at the given resolution, leading to fine-grained recognition errors. However, there remains a lack of systematic benchmarks that can serve as diagnostic tools for hallucinations in the remote sensing domain, as existing benchmarks~\cite{wang2025xlrs, LRS-VQA, LHRS, li2024vrsbench} predominantly rely on answer correctness as the primary evaluation metric.
}

{
To bridge this gap, we propose RSHBench, a benchmark for evaluating hallucinations in RS-VQA. Under a fixed generation protocol, RSHBench requires models to produce both an explicit reasoning process and a final answer, thereby standardizing generation formats across diverse architectures and reducing evaluation variance. Then, a reproducible judgment procedure is applied, leveraging a unified taxonomy to diagnose and attribute different types of hallucinations. Specifically, RSHBench categorizes grounding-related errors, such as incorrect object identification, attribute misinterpretation, and spatial relationship confusion, as Factual Hallucinations, and distinguishes them from Logical Hallucinations, which arise from invalid inference and internal inconsistencies. Based on analyses using RSHBench, we obtain several key findings:
(1) Even state-of-the-art proprietary MLLMs remain heavily affected by both factual and logical hallucinations in remote sensing scenarios.
(2) Factual hallucinations primarily stem from ungrounded object claims and attribute misidentifications, whereas logical hallucinations most often manifest as invalid reasoning and semantic over-attribution.
(3) A strong correlation exists between invalid reasoning and factual hallucinations, particularly those involving object and attribute errors, suggesting a latent snowballing effect, in which initial visual grounding failures propagate into subsequent reasoning defects.
}

{To address the above issues}, we further propose RADAR, a scalable and training-free inference framework. RADAR formulates reasoning as a two-stage, adaptive zoom-in process guided by attention that progressively refines task-relevant visual evidence. In the first stage, a location-oriented prompt guides the model to localize target regions relevant to the question, enabling adaptive and query-conditioned spatial localization. In the second stage, a content-oriented prompt is employed within these selected regions to extract fine-grained visual details and facilitate local reasoning. The final answer is derived by integrating multimodal information from both the global context and the refined local regions.

In summary, our main contributions are three-fold:

\noindent
$\bullet$ We identify that hallucinations in RS-VQA are largely driven by the inability to effectively localize and utilize task-relevant visual evidence. To this end, we introduce RSHBench, a protocol-driven benchmark that enables fine-grained, quantitative evaluation of these hallucinations.
    
$\bullet$  We propose RADAR, a scalable and training-free inference framework that leverages intrinsic attention for progressive visual evidence acquisition to facilitate adaptive and local reasoning.

{
$\bullet$ Extensive experiments on RSHBench demonstrate that RADAR consistently improves RS-VQA performance by 2\%--4\% across diverse MLLMs, while reducing both factual and logical hallucinations by approximately 10\%.
}

\section{Related Work}
\label{sec:related_work}
\subsection{MLLMs and RS-VQA}
MLLMs~\cite{hurst2024gpt, touvron2023llama, yang2025qwen3, liu2023visual, comanici2025gemini} align multimodal representations via large-scale pretraining~\cite{wang2023large} 
and instruction tuning~\cite{zhang2023instruction}, and have shown strong performance across diverse domains, from autonomous driving~\cite{cui2024survey} to medical imaging~\cite{bai2024m3d}.
Nevertheless, applying MLLMs to remote sensing remains challenging due to the distinct data distribution: remote sensing imagery is characterized by overhead perspectives, 
vast spatial extents, and a high concentration of small-scale instances.
To facilitate evaluation, recent benchmarks provide diverse remote sensing question types and reasoning requirements.
For example, LRS-VQA~\cite{LRS-VQA} targets structural reasoning on high-resolution imagery;
VRSBench~\cite{li2024vrsbench} supports captioning, RS-VQA, and localization; 
XLRS-Bench~\cite{wang2025xlrs} explores large-scale remote reasoning settings; 
MME-RealWorld-RS~\cite{Mme-realworld} emphasizes localization and attribute inference in real-world multimodal reasoning; 
and GeoChat-Bench~\cite{kuckreja2024geochat} extends evaluation toward multimodal instruction following in remote sensing.
Despite these efforts, existing benchmarks largely prioritize answer accuracy, leaving hallucination behaviors and their causes insufficiently characterized.

{
\subsection{Hallucination in Vision-Language Models}
Hallucination is a persistent failure mode of vision-language models, where generated answers or rationales are not faithfully supported by visual evidence~\cite{liu2024survey, gunjal2024detecting,wang2023evaluation}. Prior work has investigated hallucination in image captioning and VQA, and proposed both protocol-based evaluations and mitigation strategies~\cite{leng2024mitigating, zhou2023analyzing}. While hallucination has been extensively studied in general vision-language tasks, its manifestation in remote sensing scenarios remains underexplored. Moreover, existing RS-VQA evaluations rarely disentangle different hallucination types and their causes within a unified protocol, limiting systematic diagnosis and comparison. Our proposed RSHBench directly addresses this limitation by providing fine-grained, model-agnostic, and reproducible hallucination analysis.
}

{
\subsection{Attention-Based Localization and Spatial Reasoning}

Prior work has explored attention and spatial localization as key mechanisms for grounded vision-language reasoning, particularly for identifying question-relevant visual evidence.
Early studies on interpretability provide post-hoc explanations of model behavior, including attention rollout or flow~\cite{abnar-zuidema-2020-quantifying}, Grad-CAM~\cite{selvaraju2017grad}, and SmoothGrad~\cite{Smilkov2017SmoothGradRN}. 
While informative, such approaches do not explicitly guide spatial reasoning during inference. To address this limitation, region-based methods~\cite{LRS-VQA, zhou2025look, zhang2025mllms, Kang_2025_CVPR} explicitly allocate computation to selected regions to improve localization and inference. However, many of these methods rely on question-agnostic attention or generic saliency cues, which may emphasize visually prominent yet query-irrelevant regions, limiting their effectiveness in complex scenes. More recent approaches adopt coarse-to-fine localization strategies to progressively narrow down relevant regions before answering. For example, RFM~\cite{LRS-VQA} introduces a token pruning mechanism to dynamically select question-relevant image patches in gigapixel-scale imagery, while ZoomSearch~\cite{zhou2025look} employs a hierarchical zoom-in strategy to iteratively refine attended regions in ultra-high-resolution images. In contrast, RADAR leverages intrinsic attention mechanisms within MLLMs to iteratively refine visual evidence during the reasoning process, grounding both intermediate reasoning steps and final predictions in localized visual context.
}

\section{Method}
\label{sec:method}

\subsection{RSHBench: Protocol-Driven Hallucination Diagnosis in RS-VQA}
We introduce RSHBench, a protocol-driven benchmark for evaluating hallucination in RS-VQA. RSHBench comprises three key components: (i) a curated evaluation set tailored for hallucination analysis, (ii) a standardized generation protocol to ensure consistent model outputs, and (iii) a judge-based diagnosis protocol that produces fine-grained hallucination annotations.

\textbf{Evaluation set.}
We construct the evaluation set by collecting, cleaning, and sampling instances from four representative RS-VQA and remote sensing reasoning benchmarks: 
LRS-VQA~\cite{LRS-VQA}, MME-RealWorld-RS~\cite{Mme-realworld}, UCM~\cite{UCM}, and LHRS-Bench~\cite{LHRS}. 
The final dataset comprises 371 image-question pairs spanning a diverse range of task types, including structural reasoning, localization, attribute inference, and counting. 
Details on dataset composition and sampling strategy are provided in Appendix~\ref{app:data}.

\textbf{Model generation protocol.}
Given an image-question pair $(I,Q)$ (and a reference answer $A$), each evaluated model generates (i) an explicit reasoning text $C'$ and (ii) a final predicted answer $A'$ under a structured generation protocol.
This reduces format-induced variance across models and enables model-agnostic diagnosis.
Full prompts and formatting rules are provided in Appendix~\ref{app:protocol}.

\textbf{Hallucination evaluation protocol.}  
We diagnose hallucination from model outputs rather than answer accuracy.  
Hallucination is defined as any statement in $(C',A')$ not supported by the visual evidence in $I$ under the constraints of $Q$.  
We annotate at scale with three multimodal expert judges (Gemini-3-pro \cite{openai2025gpt52}, GPT-5.2 \cite{deepmind2025gemini3pro}, and Qwen3-max \cite{bai2025qwen3vltechnicalreport}) using a unified interface.  
For each tuple $(I,Q,A,C',A')$, each judge returns a JSON record with:  
(i) a binary flag \texttt{hallucinated};  
(ii) coarse categories \texttt{Factual} and \texttt{Logical};  
(iii) multi-label subtype and reason tags (Appendix~\ref{app:protocol}).  
Factual subtypes include unsupported object/category (\texttt{OBJ}), attribute (\texttt{ATT}), and spatial/location (\texttt{SPA}) claims.  
Logical subtypes include invalid reasoning (\texttt{IR}), unjustified causal inference (\texttt{CI}), internal inconsistency (\texttt{INC}), and semantic over-attribution (\texttt{SO}).  
Reference answers $A$ are used only to identify contradictions; hallucination decisions are based solely on image evidence.

\begin{figure*}[t]
    \centering
    \includegraphics[width=0.98\linewidth]{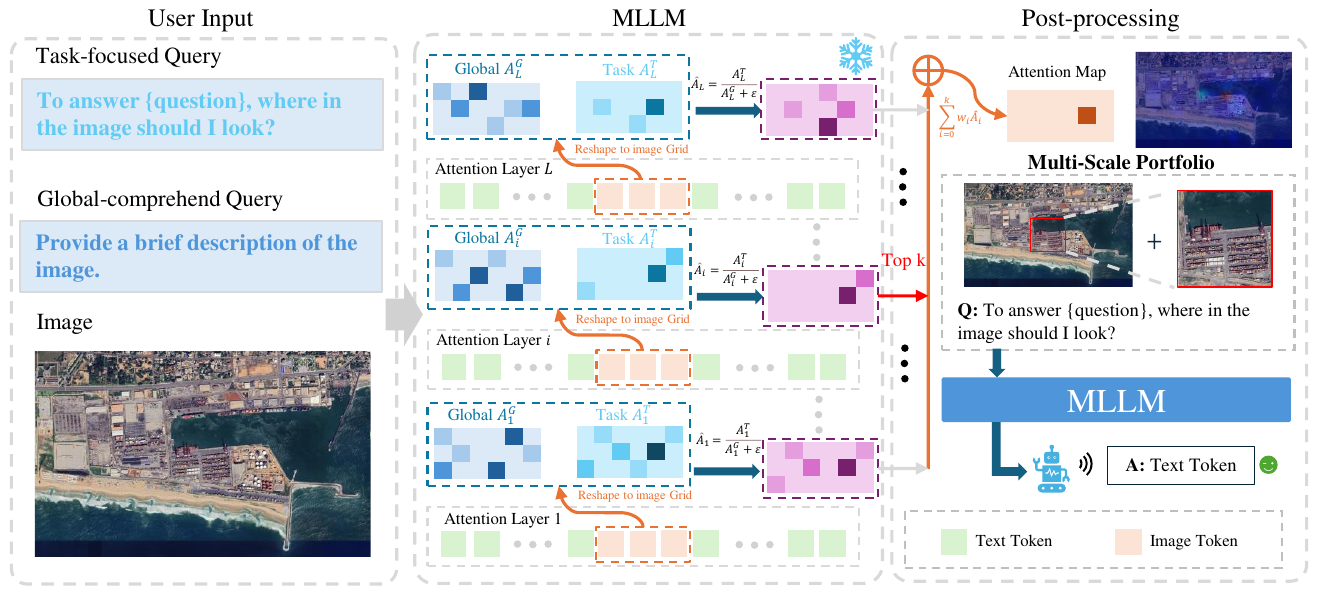} 
\caption{\textbf{Query-Conditioned Relative Attention (QCRA).}
Given an input image and two prompts: a task-focused query $Q$ (top) and a global-comprehension query $Q^{G}$ (bottom), we derive layer-wise attention relevance maps.
At each layer, token-level attention is reshaped to the image grid and a relative attention matrix is computed by contrasting the task map against the global map to suppress query-irrelevant saliency.
The most informative layers (top-$k$) are then aggregated to produce a final query-conditioned attention heatmap, which serves as the grounding signal for region selection and multi-scale evidence construction.}

\label{fig:radar}
\end{figure*}

\subsection{Relative Attention-Driven Actively Reasoning}
\label{sec:radar}
Remote sensing images are extremely large and information-dense, whereas the visual evidence required to answer a given question is often sparse and confined to small regions. 
Reasoning directly over the full image can dilute task-relevant signals and encourage reliance on linguistic priors, leading to visually unsupported predictions.
RADAR recasts visual grounding as a test-time adaptive zoom-in process to effectively mitigate hallucinations.
Before answering, the model actively localizes and progressively refines the question-relevant evidence through a coarse-to-fine procedure. 
Given an image-question pair $(I, Q)$, RADAR first performs progressive evidence acquisition and then generates the final answer via multi-view grounding.
Specifically, RADAR employs a where-oriented query to localize candidate regions that are likely to contain relevant evidence, followed by a what-oriented query to refine the evidence for fine-grained recognition. 
Both stages are driven by query-conditioned relative attention (QCRA), and regulated by a focus test that suppresses region cropping when grounding confidence is insufficient. 
Finally, the model generates the answer by synthesizing multi-view information, conditioned on both the refined visual evidence and the original image annotated with the spatial location of said evidence.
The templates used to construct $Q^{T_{\text{1}}}$ and $Q^{T_{\text{2}}}$ are predefined and detailed in Appendix~\ref{app:OPT}.

\subsubsection{Query-Conditioned Relative Attention}
RADAR is a training-free approach that exploits the internal attention patterns of MLLMs through QCRA, as illustrated in Figure~\ref{fig:radar}.
Given an image $I$ and a textual query $Q$, we extract an attention-derived spatial relevance matrix $A(Q; I)$, which reflects the image regions attended by the model when responding to $Q$. 
However, absolute attention maps are often dominated by generic visual saliency that is weakly related to the query semantics.
To suppress such query-irrelevant saliency, RADAR introduces a relative attention formulation that contrasts the task query $Q$ with a global scene query $Q^{G}$ designed to capture holistic image content. 
For each attention layer $\ell$, the relative attention matrix is defined as:
\begin{equation}
\hat{A}_{\ell}(Q; I)=\frac{A^{T}_{\ell}(Q^{T}; I)}{A^{G}_{\ell}(Q^{G}; I)+\epsilon},
\label{eq:relative_ratio}
\end{equation}
where $\ell$ is the attention layer and $\epsilon$ is a small constant for numerical stability.This ratio emphasizes regions whose attention is selectively induced by the task query rather than by global visual prominence.We compute a scalar score for each layer by summing relative attention over all spatial locations,
$s_\ell = \sum_{u} \hat{A}_\ell(u)$.
$s_\ell = \sum_{u} \hat{A}_\ell(u)$.
The top-$k$ layers with the highest scores are selected, forming the index set $\mathcal{S}_k$. For each selected layer, a normalized weight is assigned as
$w_\ell = \frac{s_\ell}{\sum_{j \in \mathcal{S}_k} s_j}$.
The final relative attention matrix is then obtained by aggregating the selected layers through a weighted sum,
\begin{equation}
\tilde{A}(Q; I)=\mathcal{N}\!\left(\sum_{\ell \in \mathcal{S}_k} w_\ell \, \hat{A}_{\ell}(Q; I)\right),
\label{eq:relative_attention}
\end{equation}
where $\mathcal{N}(\cdot)$ normalizes the result into a probability distribution over spatial cells. 
The resulting $\tilde{A}$ highlights regions whose relevance arises from query-specific amplification, rather than from generic visual saliency.

When the model fails to reliably align the question with the image, the resulting attention matrix often becomes highly diffuse, approaching a near-uniform distribution over spatial locations, as illustrated in Figure~\ref{fig:motivation}(Type1). 
Blind cropping under such uncertainty is prone to selecting spuriously relevant regions, which can further exacerbate hallucinations.
To mitigate this risk, RADAR introduces a lightweight focus test $\mathcal{F}(\cdot)$ that evaluates whether the relative attention matrix $\tilde{A}$ exhibits sufficient spatial concentration, measured for example by entropy or cumulative top-mass criteria. 
If $\mathcal{F}(\tilde{A}) \ge \tau$, RADAR activates the zoom-in mechanism to proceed with further evidence acquisition.

\subsubsection{Progressive Evidence Acquisition: Where then What}
RADAR employs an adaptive zoom-in strategy via two-stage evidence refinement that explicitly addresses the two predominant failure modes in RS-VQA, namely the inability to localize relevant evidence and the inability to recognize it with sufficient detail.
\textbf{Stage 1: Where-oriented evidence acquisition.}
We construct a location-oriented query $Q^{T_{\text{1}}}$ that prompts the model to identify image regions relevant to answering the original question $Q$. The corresponding relative attention matrix is computed as
\begin{equation}
\tilde{A}_{\text{where}} = \tilde{A}(Q^{T_{\text{1}}}; I).
\end{equation}
If $\tilde{A}_{\text{where}}$ satisfies the focus test $\mathcal{F}(\tilde{A}_{\text{where}})$, a coarse localization region is extracted using a deterministic operator $\Psi(\cdot)$, such as the tightest bounding box that covers the top-$p$ probability mass,
\begin{equation}
B_{\text{1}} = \Psi(\tilde{A}_{\text{where}}), 
\qquad
I^{\text{1}} = I[B_{\text{1}}].
\end{equation}
This stage isolates question-relevant context from large-scale remote sensing imagery and alleviates the \emph{cannot find} failure mode.

\textbf{Stage 2: What-oriented evidence refinement.}
Conditioned on the coarse crop $I^{\text{1}}$, we construct a content-oriented query $Q^{T_{\text{2}}}$ that emphasizes the information required for fine-grained recognition within the localized region. The corresponding relative attention matrix is computed as
\begin{equation}
\tilde{A}_{\text{what}} = \tilde{A}(Q^{T_{\text{2}}}; I^{\text{1}}).
\end{equation}
If $\tilde{A}_{\text{what}}$ satisfies the focus test $\mathcal{F}(\tilde{A}_{\text{what}})$, evidence is further refined by extracting a tighter region,
\begin{equation}
B_{\text{2}} = \Psi(\tilde{A}_{\text{what}}), 
\qquad
I^{\text{2}} = I^{\text{1}}[B_{\text{2}}].
\end{equation}
This stage mitigates the \emph{cannot see clearly} failure mode by increasing the effective resolution on subtle targets and suppressing local clutter, enabling more reliable recognition of attributes and object counts.

\paragraph{Multi-view grounded answering with conservative fallback.}
For answer generation, RADAR conditions the model on two complementary visual views. The first is the full image annotated with the localized region, which preserves global spatial context. The second is a high-resolution crop that supports fine-grained recognition. This design allows spatial expressions such as ``left'' or ``corner'' to be interpreted in the coordinate frame of the full image, while resolving local details from the cropped view.
To prevent overly aggressive cropping under uncertain grounding, RADAR adopts a conservative fallback strategy. If Stage1 fails the focus test, the model answers directly from the full image. If Stage1 succeeds but Stage2 fails, the model conditions on the Stage1 crop $I^{\text{1}}$. Otherwise, the refined crop $I^{\text{2}}$ is used for final answering.

\section{Experiments}
\label{sec:experiments}

\subsection{Hallucination Diagnosis on RSHBench}
\begin{table}[h]
\centering
\caption{\textbf{Leave-one-out (LOO) agreement of expert judges for the binary hallucination decision.}
We pool all evaluated base models and compare each judge against an LOO pseudo-label defined only when the other two judges agree.
We report Accuracy, Cohen's $\kappa$, and MCC .}
\label{tab:loo_similarity}
\setlength{\tabcolsep}{6pt}
\renewcommand{\arraystretch}{1.12}
\resizebox{0.82\linewidth}{!}{%
\begin{tabular}{l c c c}
\toprule
Judge                & Accuracy & Cohen's $\kappa$ & MCC \\
\midrule
Gemini-3-pro         & 0.7882 & 0.5726 & 0.5770 \\
GPT-5.2              & \textbf{0.9288} & \textbf{0.8553} & \textbf{0.8591} \\
Qwen3-max            & 0.9045 & 0.8058 & 0.8070 \\
\bottomrule
\end{tabular}%
}
\end{table}

\subsubsection{Hallucination Metrics and Judge Reliability}
We assess hallucination judgments using three standard agreement metrics(Accuracy, Cohen's $\kappa$, and MCC) to evaluate both consistency and robustness of binary hallucination decisions. 
Rather than reintroducing these metrics, we focus on their reliability in large-scale evaluation settings characterized by class imbalance and borderline cases, which are common in RS-VQA.
To quantify judge reliability without requiring human annotations at scale, we adopt a LOO agreement protocol. For each expert judge, a pseudo-label is constructed on instances where the other two judges agree. Agreement between the held-out judge and this LOO label is then measured and aggregated across all evaluated base models. This protocol yields a conservative estimate of individual judge consistency while avoiding circular effects induced by simple majority voting.

As shown in Table~\ref{tab:loo_similarity}, GPT-5.2 and Qwen3-max exhibit strong and stable agreement with the LOO labels (Accuracy $> 0.90$, $\kappa > 0.80$, MCC $> 0.80$), indicating highly consistent hallucination judgments across models. In contrast, Gemini-3-pro shows lower yet meaningful agreement ($\kappa = 0.57$), suggesting greater sensitivity to borderline cases such as weak visual evidence or ambiguous grounding. This variability motivates the use of multi-judge consensus rather than reliance on a single evaluator.
To further validate alignment with human perception, we conduct human spot-check calibration on a randomly sampled subset. As detailed in Appendix~\ref{app:agreement}, the relative ranking of judges remains consistent when compared with human annotations, supporting the reliability of the consensus-based hallucination labels used throughout our experiments.

\begin{table}[h]
\small
\centering
\caption{\textbf{Hallucination evaluation on RSHBench.}
Each model prediction is annotated by three expert judges with a binary hallucination rate ($HR$) and a multi-label taxonomy comprising factual subtypes (\texttt{OBJ}, \texttt{ATT}, \texttt{SPA}) and logical subtypes (\texttt{IR}, \texttt{CI}, \texttt{INC}, \texttt{SO}), along with coarse categories $HR_{F}$ (Factual) and $HR_{L}$ (Logical).
The overall $HR$ is determined by majority vote, while category and subtype labels are aggregated by taking the union of tags from judges who marked the instance as {hallucinated}.
All values are reported as percentages over the full evaluation set .}
\label{tab:hallu_consensus}

\resizebox{\linewidth}{!}{%
\begin{tabular}{
l
c c c >{\columncolor{blue!15}}c
c c c c >{\columncolor{blue!15}}c
>{\columncolor{blue!15}}c
}
\toprule
\multirow{2}{*}{Models} &
\multicolumn{4}{c}{Factual Hallucination} &
\multicolumn{5}{c}{Logical Hallucination} &
\multicolumn{1}{>{\columncolor{blue!15}}c}{\textbf{HR}} \\
\cmidrule(lr){2-5}\cmidrule(lr){6-10}
& OBJ & ATT & SPA & $HR_F$ & IR & CI & INC & SO & $HR_L$
& \multicolumn{1}{>{\columncolor{blue!15}}c}{} \\

\midrule
\multicolumn{11}{l}{\textit{\textcolor{gray}{Closed-source Models}}} \\
Claude-3-7      & 40.70 & 28.30 & 11.32 & 55.53 & 20.49 & 0.27 & 1.08 & 14.29 & 24.53 & 56.33 \\
Gemini-2.5-pro  & 33.42 & 31.54 & 15.36 & 48.79 & 27.49 & 0.54 & 0.81 & 15.09 & 29.65 & 49.06 \\
GPT-4o          & 30.19 & 26.68 & 11.32 & 46.90 & \textbf{19.41} & 0.27 & \textbf{0.27} & 11.05 & \textbf{21.02} & 47.44 \\
\midrule
\multicolumn{11}{l}{\textit{\textcolor{gray}{Open-source Models}}} \\
GLM-4.6v        & 35.31 & \textbf{21.02} & \textbf{9.16} & 48.79 & 21.02 & \textbf{0.00} & 0.54 & \textbf{9.16} & 22.91 & 49.60 \\
LLaVA-1.5-7B    & \textbf{25.88} & 29.11 & 14.29 & 46.90 & 25.61 & 2.96 & 2.70 & 16.71 & 26.95 & 47.71 \\
Qwen3-VL-4B     & 44.47 & 31.81 & 14.82 & 61.19 & 29.92 & 0.27 & 2.43 & 15.63 & 34.77 & 61.19 \\
LLaMA-3.2-90B   & 35.85 & 25.88 & 12.13 & 51.48 & 26.15 & 0.27 & 3.77 & 15.36 & 29.11 & 52.02 \\
GeoZero         & 33.42 & 33.96 & 15.90 & 49.87 & 28.30 & 3.77 & 2.16 & 18.06 & 29.65 & 49.87 \\
\midrule
GeoZero\textbf{+RADAR} &
28.03 & 25.61 & 13.48 & \textbf{38.54} &
21.83 & 2.16 & 1.89 & 15.63 & 24.80 & \textbf{38.81} \\
\bottomrule
\end{tabular}%
}
\end{table}
\subsubsection{Hallucination Evaluation Results}
Table~\ref{tab:hallu_consensus} shows that hallucinations are prevalent across both closed- and open-source MLLMs, with most baselines exhibiting a hallucination rate ($HR$) between 47\% and 61\%. 
Factual hallucinations are the most dominant, with \texttt{OBJ} and \texttt{ATT} consistently comprising the largest proportions. These reflect frequent confusions in object/category recognition and attribute identification, particularly when visual evidence is subtle or fine-grained. Spatial errors (\texttt{SPA}) are also substantial, highlighting persistent challenges in grounding predictions to the correct regions referenced by the question.
On the {logical} dimension, \texttt{IR} and \texttt{SO} occur more frequently than direct \texttt{CI}, suggesting that hallucinations often stem from flawed reasoning or internal inconsistency rather than explicit disagreement with the reference answer. Due to the multi-label annotation design, logical tags frequently co-occur with factual ones (e.g., \texttt{IR} with \texttt{OBJ}/\texttt{ATT}), indicating compounded failure patterns where incorrect grounding cascades into faulty inference.

We next compare RADAR with a diverse set of strong baselines, including both general-purpose MLLMs and remote-sensing–specialized models. Across all evaluated backbones, RADAR consistently reduces hallucination rates. 
Compared with strong RS-specialized baselines, our approach significantly reduces the overall hallucination rate to $HR = 38.81\%$. Specifically, it lowers $HR$ by 11.06 percentage points relative to GeoZero (from 49.87\%). Most of the improvement is concentrated in factual subtypes (\texttt{OBJ}/\texttt{ATT}), which aligns with two core bottlenecks in RS-VQA:  
(i) failures to localize question-relevant evidence in large-scale imagery (reflected in \texttt{SPA}/\texttt{OBJ}), and  
(ii) failures to extract fine-grained semantics after localization (reflected in \texttt{ATT}/\texttt{OBJ}).
These findings underscore the importance of enhancing question-conditioned localization and fine-grained evidence extraction.

\subsection{Effectiveness of RADAR}

\subsubsection{Experimental Settings}

We evaluate RADAR on three representative benchmarks that collectively cover fine-grained recognition, spatial grounding, and reasoning in remote sensing: LRS-VQA~\cite{LRS-VQA}, MME-RealWorld-RS~\cite{Mme-realworld}, and LHRS-Bench~\cite{LHRS}. 
LRS-VQA can effectively assess the large remote sensing imagery perception capability of LVLMs. 
MME-RealWorld-RS focuses on real-world scenarios that require precise localization and attribute-level discrimination under challenging visual conditions. 
LHRS-Bench adopts a structured protocol that combines label filtering, balanced sampling, and LLM-generated instruction prompts to construct tasks spanning recognition, spatial perception, and reasoning.
We report results across a diverse set of baselines, including both general-purpose and remote-sensing-specialized multimodal large language models. 
Specifically, we consider 
(i) \emph{closed-source MLLMs}, including GPT-4o~\cite{hurst2024gpt}, Gemini-2.5-Pro~\cite{comanici2025gemini}, and Claude-4.5 (Haiku variant); 
(ii) \emph{open-source MLLMs}, such as LLaVA-v1.5~\cite{liu2023visual}, Qwen-VL~\cite{yang2025qwen3}, LLaMA~\cite{touvron2023llama}, and GLM (4.6v and 4.5v); 
and (iii) \emph{remote-sensing-oriented models}, including GeoChat~\cite{kuckreja2024geochat}, and GeoZero~\cite{wang2025geozero}, which are explicitly designed for geospatial imagery. 
In addition, we include \emph{plug-and-play region-based inference baselines} by applying ViCrop (rel-att)~\cite{zhang2025mllms} to representative backbones, enabling a direct comparison with RADAR under the same inference-time setting. ViCrop directly computes relative attention for each image–question pair and uses the resulting attention map as an importance signal for visual cropping.

\begin{table*}[t]
\small
\centering
\caption{\textbf{Overall performance on remote sensing VQA and real-world reasoning.}
We report accuracy (\%) on LRS-VQA (FAIR, Bridge, STAR, and their average accuracy (AA)), MME-RealWorld-RS (Position, Color, Count, and AA), and LHRS-Bench. \textbf{Avg.} denotes the mean score across the three benchmarks (adopting AA for LRS-VQA and MME-RealWorld-RS). \textbf{RADAR} indicates the application of our training-free inference procedure to the GeoZero backbone (see Table~\ref{tab:farsight_main} for per-backbone gains), achieving the highest average performance among all evaluated methods.}

\label{tab:main_results}
\resizebox{0.98\linewidth}{!}{
\begin{tabular}{lcccccccccc}
\toprule
\multirow{2}{*}{\textbf{Methods}}
& \multicolumn{4}{c}{\textbf{LRS-VQA}}
& \multicolumn{4}{c}{\textbf{MME-RealWorld-RS}}
& \textbf{LHRS-Bench}
& \multirow{2}{*}{\textbf{Avg.}} \\
\cmidrule(lr){2-5} \cmidrule(lr){6-9} \cmidrule(lr){10-10}
& FAIR & Bridge & STAR & AA
& Position & Color & Count & AA
& -- &  \\
\midrule

Llama-4-scout        & 19.72 & 29.19 & 23.73 & 24.21 & 26.70 & 23.72 & 15.64 & 22.02 & 37.33 & 27.85 \\
Claude-haiku-4-5     &  5.68 &  9.79 &  2.40 &  5.96 & 31.73 & 28.04 & 18.35 & 26.04 & 55.01 & 29.00 \\
GLM-4.6v             & 27.55 & 27.31 & 29.31 & 28.06 & 44.92 & 28.19 &  7.53 & 26.88 & 67.19 & 40.71 \\
GPT-4o               & 22.89 & 24.39 & 29.78 & 25.69 & 36.37 & 32.43 & 15.89 & 28.23 & 66.19 & 40.04 \\
\midrule

Qwen2.5-VL-7B       & 22.18 & 27.40 & 26.71 & 25.43 & 46.22 & 29.16 & 11.26 & 28.88 & 53.98 & 36.10 \\
Qwen3-VL-4B         & 26.23 & 29.47 & 32.86 & 29.52 & 55.53 & 40.24 &  7.59 & 34.45 & 65.35 & 43.11 \\
Qwen3-VL-8B         & 29.71 & 32.77 & \textbf{35.01} & 32.49 & 54.97 & 46.06 & 14.44 & 38.49 & 66.03 & 45.67 \\
LLaVA-1.5-7B        & 20.03 & 27.97 & 26.96 & 24.99 & 33.73 & 35.78 & 18.52 & 29.34 & 47.17 & 33.83 \\
\midrule

GeoChat            & 20.18 & 24.54 & 13.75 & 19.49 & 25.06 & 23.11 & 15.66 & 21.28 & 37.62 & 26.13 \\
GeoZero            & 29.53 & 31.26 & 33.96 & 31.58 & 57.04 & 44.30 & 15.74 & 39.03 & 66.08 & 45.56 \\
\midrule

\rowcolor{gray!15}
\textbf{RADAR}
& \textbf{31.21} 
& \textbf{33.33}
& 34.11
& \textbf{32.88}
& \textbf{58.15} 
& \textbf{50.52}
& \textbf{20.47} 
& \textbf{43.05}
& \textbf{67.47}
& \textbf{47.40}  \\

\bottomrule
\end{tabular}
}
\end{table*}

\begin{table*}[ht] 
\centering
\caption{\textbf{Effect of evidence cropping and progressive localization.}
We compare each backbone model with a generic cropping baseline (\textbf{+ViCrop}) and our (\textbf{+RADAR}). Reported values are accuracy (\%) on each subset, and red numbers indicate absolute improvements over the corresponding backbone. ViCrop yields limited and often unstable gains, whereas RADAR consistently improves performance, particularly on attribute recognition and counting tasks that require fine-grained visual evidence.}

\label{tab:farsight_main}
\resizebox{0.98\linewidth}{!}{
\begin{tabular}{l|ccc|ccc|c}
\toprule
\multirow{2}{*}{\textbf{Method}} 
& \multicolumn{3}{c|}{\textbf{LRS-VQA}}
& \multicolumn{3}{c|}{\textbf{MME-RealWorld-RS}}
& \multicolumn{1}{c}{\textbf{LHRS-Bench}} \\
\cmidrule(lr){2-4} \cmidrule(lr){5-7} \cmidrule(lr){8-8}
& FAIR $\uparrow$ & Bridge $\uparrow$ & STAR $\uparrow$
& Position $\uparrow$ & Color $\uparrow$& Count $\uparrow$ 
& -- \\
\midrule

LLaVA-OV
& --    & --    & --   
& 18.70 & 14.26 & 17.37
& 19.66 \\
\rowcolor{gray!15}
+ \textbf{RADAR (Ours)}
& --    & --    & --   
& 22.83 {\color{red}{(+4.14)}} & 32.83 {\color{red}{(+18.57)}}& \textbf{29.53} {\color{red}{(+12.15)}} 
& 23.03 {\color{red}{(+3.37)}} \\
\midrule

Qwen3-VL
& 26.23 & 29.47 & 32.86 
& 55.53 & 40.24 &  7.59 
& 65.35\\
+ ViCrop
& 27.46 & 28.91 & 33.31
& 54.57 & 42.87 & 10.60 
& 63.51 \\
\rowcolor{gray!15}
+ \textbf{RADAR (Ours)}
& 30.77 {\color{red}{(+4.54)}}
& 29.94 {\color{red}{(+0.47)}}
& \textbf{34.98} {\color{red}{(+2.13)}}
& 56.64 {\color{red}{(+1.11)}}
& \textbf{53.71} {\color{red}{(+13.47)}}
& 15.01 {\color{red}{(+7.42)}}
& \textbf{67.73} {\color{red}{(+2.38)}}\\
\midrule

GeoZero
& 29.53 & 31.26 & 33.96
& 57.04 & 44.30 & 15.74 
& 66.08 \\
+ ViCrop
& 28.83 & 31.73 & 32.91
& 57.04 & 47.41 & 18.19 
& 65.14 \\
\rowcolor{gray!15}
+ \textbf{RADAR (Ours)}
& \textbf{31.21} {\color{red}{(+1.68)}}
& \textbf{33.33} {\color{red}{(+2.07)}}
& 34.11 {\color{red}{(+0.15)}}
& \textbf{58.15} {\color{red}{(+1.11)}}
& 50.52 {\color{red}{(+6.22)}}
& {20.47} {\color{red}{(+4.73)}}
& 67.47 {\color{red}{(+1.39)}} \\
\bottomrule
\end{tabular}
}
\end{table*}

\subsubsection{Quantitative Results}
\label{sec:quant_results}
\begin{figure*}[h]

    \centering

    \includegraphics[width=1\linewidth]{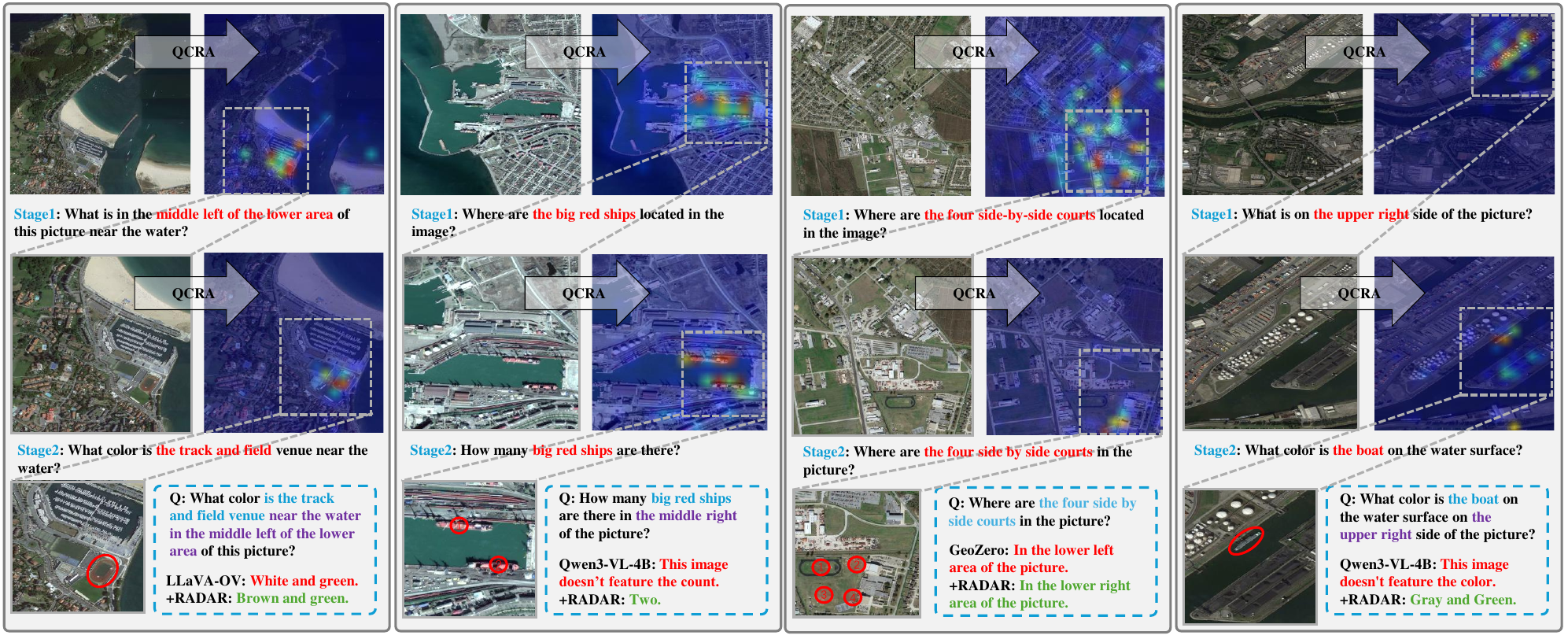} 
\caption{\textbf{Qualitative examples of RADAR's QCRA and progressive evidence refinement.}
For each example, we visualize the QCRA heatmap generated by a \emph{where}-oriented query on the full image (top) and a \emph{what}-oriented query on the localized crop (middle), where brighter regions indicate stronger query-conditioned relevance. Dashed boxes mark the regions selected for zoom-in evidence extraction.}
\vspace{-0.2cm}
    \label{fig:qualitative}

\end{figure*}

Table~\ref{tab:main_results} summarizes the results on three representative benchmarks. Applying RADAR to the GeoZero backbone achieves the highest average performance among all evaluated methods, improving \textbf{Avg.} from 45.56 to 47.40 (+1.84) without additional training. Notably, RADAR also outperforms the strongest direct baseline, Qwen3-VL-8B, which attains an average score of 45.67, by a margin of 1.73 points. This result indicates that test-time evidence acquisition can be competitive with scaling the backbone model.
On LRS-VQA, RADAR improves GeoZero from 31.58 to 32.88 (+1.30) in average accuracy, with gains on both FAIR (+1.68) and Bridge (+2.07). These improvements suggest more effective question-conditioned localization in large-scale scenes. On LHRS-Bench, RADAR yields a consistent improvement from 66.08 to 67.47 (+1.39), indicating that local evidence refinement remains beneficial even when global recognition performance is relatively strong.
The largest gains are observed on MME-RealWorld-RS, where GeoZero improves from 39.03 to \textbf{43.05} (+4.02) in average accuracy. In particular, substantial improvements are achieved on Color (+6.22) and Count (+4.73), supporting the hypothesis that many RS-VQA failures stem from insufficient fine-grained visual evidence rather than from missing global context.

Table~\ref{tab:farsight_main} isolates the effect of visual evidence cropping. A generic cropping strategy ({+ViCrop}) yields limited and sometimes negative gains. For instance, it reduces LHRS-Bench accuracy for both Qwen3-VL (65.35 to 63.51) and GeoZero (66.08 to 65.14), indicating that unguided cropping can discard relevant context or select mismatched regions. In contrast, RADAR consistently improves performance across backbones and subsets, demonstrating the importance of query-conditioned localization and progressive refinement.
On Qwen3-VL, RADAR improves LRS-VQA FAIR by +4.54 (26.23 to 30.77) and increases MME-RealWorld-RS Color by +13.47 (40.24 to 53.71), whereas {+ViCrop} yields only a +2.63 gain on Color and degrades performance on Position. Similarly, on GeoZero, RADAR improves all three MME-RealWorld-RS subsets, including Count (+4.73), while {+ViCrop} provides smaller and less stable gains across LRS-VQA subsets. These results indicate that RADAR benefits not merely from cropping, but from how regions are selected and progressively refined.
Overall, the results in Tables~\ref{tab:main_results}-~\ref{tab:farsight_main} show that RADAR improves accuracy without sacrificing faithfulness by enforcing question-relevant evidence acquisition prior to prediction.

\subsubsection{Qualitative results}
Figure~\ref{fig:qualitative} illustrates QCRA map and the resulting evidence refinement on representative failure cases. When questions depend on sparse and small-scale visual evidence, baseline models often produce diffuse or misaligned attention maps, leading to reliance on priors or weakly supported answers. In contrast, RADAR highlights regions that are directly relevant to the query and enables more reliable extraction of local evidence.
For instance, RADAR corrects a fine-grained attribute prediction by zooming into the track-and-field venue near the shoreline, changing the response from ``White and green'' (LLaVA-OV) to ``Brown and green''. It recovers counting by focusing on the queried ships, correcting Qwen3-VL’s failure to provide a count to the accurate answer ``Two''. RADAR also resolves a spatial grounding error by localizing the correct court region, revising GeoZero’s prediction from ``lower left'' to ``lower right''. These examples demonstrate that improved question-conditioned localization and local inspection reduce visually unsupported claims and lead to more grounded predictions.

\begin{table}[h]
\centering
\caption{Ablation study on the operation stages of RADAR, conducted with the Qwen3-VL-4B model. We report accuracy on MME-RealWorld-RS and LHRS-Bench.}
\label{tab:visual_funnel_ablation}
\resizebox{0.92\linewidth}{!}{
\begin{tabular}{lcc}
\toprule
\textbf{Configuration} & \textbf{MME-RealWorld} & \textbf{LHRS-Bench} \\
\midrule
Baseline         & 34.45 & 65.36 \\
RADAR w/o Stage 2 & 39.05 & 66.17 \\
RADAR w/o Stage 1 & 38.88 & 66.87 \\
\hline
\rowcolor{gray!15}
\textbf{RADAR}   & \textbf{41.79} & \textbf{67.73} \\
\bottomrule
\end{tabular}}
\end{table}

\subsubsection{Ablation study}
Table~\ref{tab:visual_funnel_ablation} presents an ablation of the two-stage evidence refinement strategy in RADAR using Qwen3-VL-4B. Removing either stage still improves performance over the baseline, indicating that coarse localization (Stage1) and local refinement (Stage2) provide complementary benefits. Using only a single stage yields consistent gains on both MME-RealWorld-RS and LHRS-Bench, with improvements of up to +4.60 and +1.51 absolute points, respectively.
These results suggest that Stage1 reduces attention diffusion by isolating question-relevant context, while Stage2 further enhances fine-grained recognition within the localized region. Finally, jointly applying both components yields the most reliable and consistent performance gains.
\vspace{-0.1cm}
\section{Conclusion}
\label{sec:conclusion}

We investigate hallucination in RS-VQA and identify two dominant grounding failure modes: \emph{cannot find}, 
which refers to the inability to localize question-relevant regions in large-scale scenes, and \emph{cannot see clearly}, 
which reflects difficulty in extracting fine-grained semantics from small or weak visual evidence. To enable systematic analysis, 
we introduce \textbf{RSHBench}, a protocol-driven benchmark that provides standardized data generation and fine-grained, judge-based hallucination annotations.

Building on these insights, we propose \textbf{RADAR}, a training-free inference method that leverages query-conditioned relative attention 
to perform coarse-to-fine evidence acquisition. 
Extensive experiments across multiple MLLMs demonstrate that RADAR consistently improves RS-VQA accuracy while substantially reducing hallucinations, underscoring the effectiveness of test-time evidence acquisition for reliable remote sensing reasoning.

\vspace{-0.1cm}
\section{Limitations}
RADAR assumes access to informative internal attention signals and predefined \emph{where}/\emph{what} query templates, 
which may limit its direct applicability to fully black-box models. 
The multi-stage evidence acquisition procedure also incurs additional inference cost, 
and relative attention may remain unreliable in highly ambiguous scenes. 
RSHBench relies on MLLM-based judges and consensus labeling; 
although agreement checks and spot audits are employed, 
borderline cases may still be imperfectly annotated.

\nocite{langley00}

\section*{Impact Statement}
This paper presents work whose goal is to advance the field of Machine
Learning. There are many potential societal consequences of our work, none
which we feel must be specifically highlighted here.

\bibliography{example_paper}
\bibliographystyle{icml2026}
\newpage
\appendix
\onecolumn
\appendix

\section{Appendix Material Overview}
This Appendix material is organized into two parts. Part~I provides additional details on \textbf{RSHBench}, including benchmark curation, standardized generation and evaluation protocols, the hallucination taxonomy, and reliability analyses. Part~II presents further implementation details of \textbf{RADAR}, covering QCRA, the focus test and region extraction, and the \emph{where}/\emph{what} prompt templates.

\section{RSHBench: Benchmark Construction and Hallucination Diagnosis}
\label{app:rshbench}

\subsection{Stratified Sampling Configuration}
\label{app:data}
We construct RSHBench using stratified sampling to ensure balanced coverage across key capability axes. 
Specifically, we consider common VQA capabilities including \emph{color}, \emph{shape}, \emph{quantity}, \emph{position}, and \emph{multi-step reasoning}, 
alongside remote-sensing-specific factors including \emph{orientation}, \emph{distance}, and \emph{scene type}. 
Samples are drawn from \textit{MME-RealWorld-RS}, \textit{LRS-VQA}, and \textit{LHRS-Bench} according to per-axis quotas. 
This design prevents over-representation of any single capability and enables fine-grained analysis of hallucination patterns across diverse question types.

Before finalizing the evaluation set, we perform data cleaning to enhance label reliability. 
Duplicate items are removed, and samples with evident label errors or inconsistent annotations are discarded. 
Potential issues are identified by cross-checking the question, reference answer, and corresponding visual evidence. This procedure reduces evaluation noise and strengthens the reliability of subsequent hallucination analysis.

The final curated remote sensing image–question set consists of \textbf{371} instances. Each instance includes a remote sensing image, a question, the corresponding reference answer, and associated visual evidence. This set forms the basis for hallucination diagnosis and model evaluation throughout the paper.

\subsection{Inference Protocol and Structured Judge Interface}
\label{app:protocol}

\subsubsection{Model generation protocol}
For each instance, the evaluated model is instructed to produce an explicit reasoning trace along with a final answer in a strict JSON format. The instruction mandates that each reasoning step be grounded in observable visual evidence and explicitly prohibits guessing when evidence is insufficient.

\begin{promptbox}{Generation Protocol}
\begin{PromptVerbatim}
You are a multimodal large language model performing remote sensing visual question answering.
You will be given:
1. A remote sensing image.
2. A question about the image.
Your task is to answer the question by reasoning strictly based on the visual content of the image.

Reasoning and Answering Guidelines
- You must first provide an explicit reasoning process explaining how you derive the answer.
- Every step in your reasoning must be grounded in observable visual evidence from the image.
- Do NOT rely on external knowledge, assumptions, or common sense beyond what is visually observable.
- If the question cannot be answered with certainty from the image alone,
  you must explicitly state that the information cannot be determined from the image.

Important Constraints
- Do NOT speculate or guess.
- Do NOT introduce semantic attributes, functions, or intentions unless they are directly observable.
- Do NOT omit the reasoning process, even if the answer seems obvious.
- If visual evidence is insufficient, your reasoning should explicitly state this.

Output Format (STRICT)
You MUST output your response in the following JSON format:
{
  "reasoning": "A step-by-step explanation grounded strictly in visual evidence.",
  "answer": "The final answer to the question."
}

Question:
{question}
\end{PromptVerbatim}
\end{promptbox}

\subsubsection{Hallucination evaluation protocol}
Each expert judge receives the image $I$, question $Q$, the reference answer $A$, the model reasoning $C'$, and the predicted answer $A'$.
The judge determines whether hallucination occurs, assigns coarse categories, and returns fine-grained subtype labels with explicit evidence grounding.

\begin{promptbox}{Hallucination Evaluation Protocol (Judge)}
\begin{PromptVerbatim}
You are an expert hallucination evaluation judge for multimodal large language models (MLLMs), specialized in remote sensing visual question answering.
Your task is NOT to answer the question. Your task is to evaluate whether the model's output exhibits hallucination, based strictly on visual evidence and the provided information.
You will be given:
1. An image (I).
2. A question (Q).
3. A ground-truth answer (A), when provided.
4. The model's reasoning process (C').
5. The model's final answer (A').

You must determine:
- Whether hallucination occurs.
- Whether the hallucination is factual or logical in nature.
- The specific hallucination subtype(s).
- The underlying reason(s) or trigger(s).

Definition of Hallucination
A hallucination occurs when the model's reasoning (C') or final answer (A') contains information that is not faithfully supported by the image or by valid logical inference grounded in the image.

Hallucination is categorized into two high-level types:
1. Factual Hallucination:
Errors where the generated content contradicts, fabricates, or overstates factual information that should be directly grounded in visual evidence.
2. Logical Hallucination:
Errors where the generated content involves flawed reasoning, unjustified inference, or internal inconsistency.
An incorrect answer alone does NOT necessarily imply hallucination.

Allowed and Disallowed Information
You may ONLY use:
- The image (I).
- The question (Q).
- The ground-truth answer (A), when provided.
- The model's reasoning (C').
- The model's final answer (A').

You must NOT:
- Introduce external knowledge beyond the image.
- Re-answer the question yourself.
- Override the ground-truth answer.

Hallucination Taxonomy
If hallucination occurs, classify it into one or more subtypes.
A. Factual Hallucination
A1. Object / Category Hallucination
A2. Attribute Hallucination
A3. Spatial / Relational Hallucination

B. Logical Hallucination
B1. Invalid Reasoning
B2. Unjustified Causal Inference
B3. Internal Inconsistency
B4. Semantic Over-Attribution

Evidence Grounding Requirement
For every judgement, you MUST cite relevant visual evidence from the image, OR state that such evidence does not exist.

Output Format (STRICT JSON)
{
  "hallucinated": true/false,
  "hallucination_category": ["Factual", "Logical"],
  "hallucination_types": [],
  "hallucination_reasons": [],
  "evidence_basis": [],
  "justification": "Concise explanation grounded explicitly in visual evidence or its absence."
}
\end{PromptVerbatim}
\end{promptbox}

\begin{table}[h]
  \centering
  \caption{\textbf{Hallucination taxonomy.} Multiple labels may apply to the same prediction.}
  \label{tab:taxonomy_def}
  \setlength{\tabcolsep}{4pt}
  \renewcommand{\arraystretch}{1.08}

  \begin{tabularx}{\linewidth}{@{} c l >{\RaggedRight\arraybackslash}X @{}}
    \toprule
    \textbf{Code} & \textbf{Hallucination Types} & \textbf{Definition} \\
    \midrule
    OBJ & Object/Category & Claims the presence/absence of an object or category not supported by the image. \\
    ATT & Attribute & Misstates attributes (e.g., color, shape, size, count, material) not supported by the image. \\
    SPA & Spatial/Relational & Misstates spatial relations (e.g., relative position, containment, adjacency) between objects/regions. \\
    IR & Invalid Reasoning & Uses incorrect premises, unsupported assumptions, or invalid inference steps. \\
    CI & Unjustified Causality & Infers causality, intent, or functionality from correlation without evidence. \\
    INC & Internal Inconsistency & The reasoning process $C'$ conflicts with the final answer $A'$. \\
    SO & Semantic Over-Attribution & Assigns semantic roles/functions that cannot be determined from visual evidence. \\
    \bottomrule
  \end{tabularx}
\end{table}

\subsection{Agreement Metrics}
\label{app:Metrics}
We measure agreement for the binary hallucination label using Accuracy, Cohen's $\kappa$, and Matthews correlation coefficient (MCC).
For two labelers (or a labeler vs.\ a reference), we denote the confusion-matrix counts as:
$TP$,
$FP$,
and $FN$ .
Let $N = TP + TN + FP + FN$.

\textbf{Accuracy.}
\begin{equation}
\mathrm{Acc}=\frac{TP+TN}{N}.
\end{equation}
Accuracy reports the overall fraction of matching decisions. It is easy to interpret, but may be overly optimistic when the class distribution is imbalanced.

\textbf{Cohen's $\kappa$.}
Cohen's $\kappa$ quantifies agreement \emph{beyond chance}. It compares the observed agreement $p_o$ with the chance agreement $p_e$ implied by the two labelers' marginal label frequencies:
\begin{equation}
\kappa=\frac{p_o-p_e}{1-p_e},
\qquad
p_o=\frac{TP+TN}{N}.
\end{equation}
To compute $p_e$, let the positive/negative rates of labeler A be
$p_A^+ = \frac{TP+FP}{N}$ and $p_A^- = 1-p_A^+$,
and those of labeler B be
$p_B^+ = \frac{TP+FN}{N}$ and $p_B^- = 1-p_B^+$.
Then
\begin{equation}
p_e = p_A^+ p_B^+ + p_A^- p_B^-.
\end{equation}
Intuitively, $\kappa$ discounts agreement that would be expected even if both labelers guess according to their own label biases.

\textbf{MCC.}
\begin{equation}
\mathrm{MCC}=\frac{TP\cdot TN-FP\cdot FN}{\sqrt{(TP+FP)(TP+FN)(TN+FP)(TN+FN)}}.
\end{equation}
MCC can be viewed as a correlation between two binary labelings and summarizes all four confusion-matrix terms.
It is typically more reliable than Accuracy under class imbalance because it penalizes degenerate predictions and remains informative even when one class is rare.

\textbf{Leave-one-out agreement.}
To estimate judge reliability without requiring additional human annotations, we adopt a leave-one-out (LOO) agreement protocol over the three expert judges. Let $\{J_1, J_2, J_3\}$ denote the judges, and let $y_i^{(j)} \in \{0,1\}$ represent the binary hallucination decision of judge $J_j$ on instance $i$, where $1$ indicates a hallucinated response. For each held-out judge $J_j$, we construct an LOO pseudo label $\tilde{y}_i^{(-j)}$ only for instances on which the remaining two judges agree:
\begin{equation}
\tilde{y}_i^{(-j)}=
\begin{cases}
y_i^{(k)}, & \text{if } y_i^{(k)} = y_i^{(\ell)},\ \{k,\ell\}=\{1,2,3\}\setminus\{j\},\\
\text{undefined}, & \text{otherwise}.
\end{cases}
\end{equation}
We then compare the decisions of $J_j$ with $\tilde{y}^{(-j)}$ using Accuracy, Cohen’s $\kappa$, and MCC, computed over the subset of instances for which $\tilde{y}^{(-j)}$ is defined. To improve statistical stability, we pool these subsets across all evaluated base models, treating each model–instance pair as an evaluation item under the same protocol, and report the resulting LOO agreement scores in Table~\ref{tab:loo_similarity}. This protocol is conservative, as it evaluates each judge only on items where the other two judges reach consensus and avoids inflating agreement on inherently ambiguous cases where disagreement persists.

\subsection{Inter-Rater Reliability and Human Validation}

\label{app:agreement}
We report inter-judge reliability (IRR) in two settings.
First, we measure the agreement between each judge and the three-judge majority vote on the full evaluation set.
Second, we conduct a human calibration study on a random subset and evaluate both individual judges and the majority vote against human annotations.

\begin{table}[h]
\centering
\caption{\textbf{Agreement between each judge and the three-judge majority vote.}
We report Accuracy, Cohen's $\kappa$, and MCC.}
\label{tab:overall_similarity_majority}
\setlength{\tabcolsep}{7pt}
\renewcommand{\arraystretch}{1.12}
\resizebox{0.4\linewidth}{!}{%
\begin{tabular}{l c c c}
\toprule
Rater & Accuracy & $\kappa$ & MCC \\
\midrule
Gemini-3-pro & 0.8113 & 0.6225 & 0.6259 \\
GPT-5.2      & 0.9476 & 0.8951 & 0.8972 \\
Qwen3-max    & 0.9314 & 0.8628 & 0.8632 \\
\bottomrule
\end{tabular}
}
\end{table}

\begin{table}[h]
\centering
\caption{\textbf{Agreement between expert judges and human annotations.}
We report Accuracy, Cohen's $\kappa$, and MCC.}
\label{tab:judges_vs_human}
\setlength{\tabcolsep}{7pt}
\renewcommand{\arraystretch}{1.12}
\resizebox{0.4\linewidth}{!}{%
\begin{tabular}{l c c c}
\toprule
Rater & Accuracy & $\kappa$ & MCC \\
\midrule
Gemini-3-pro & 0.6700 & 0.3159 & 0.3165 \\
GPT-5.2      & 0.9500 & 0.8934 & 0.8953 \\
Qwen3-max    & 0.7500 & 0.5000 & 0.5126 \\
\bottomrule
\end{tabular}
}
\end{table}

\subsubsection{Inter-rater reliability and human calibration}
We first report agreement between each judge and the three-judge majority vote on the pooled evaluation set (Table~\ref{tab:overall_similarity_majority}). We further conduct a human calibration study by randomly sampling 100 instances and collecting human binary hallucination labels, and then evaluate each judge against human annotations (Table~\ref{tab:judges_vs_human}).

The results indicate strong and consistent judge reliability, with a stable ranking across both agreement settings. When compared against the three-judge majority vote (Table~\ref{tab:overall_similarity_majority}), GPT-5.2 achieves the highest agreement (Accuracy 0.9476 / $\kappa$ 0.8951 / MCC 0.8972), followed by Qwen3-max (0.9314 / 0.8628 / 0.8632), while Gemini-3-pro is substantially lower (0.8113 / 0.6225 / 0.6259). This suggests that GPT-5.2 aligns most closely with the consensus and yields the most stable binary hallucination decisions.

Agreement with human annotations (Table~\ref{tab:judges_vs_human}) further validates this trend: GPT-5.2 remains highly consistent with human labels (Accuracy 0.9500 / $\kappa$ 0.8934 / MCC 0.8953), Qwen3-max shows moderate alignment (0.7500 / 0.5000 / 0.5126), and Gemini-3-pro exhibits noticeably weaker agreement (0.6700 / 0.3159 / 0.3165). Notably, the judge–majority agreements are generally higher than judge–human agreements, which is expected given the presence of borderline cases in RS-VQA hallucination diagnosis (e.g., weak evidence, ambiguous regions), where subjective thresholds can differ. Importantly, the consistently high $\kappa$ and MCC for GPT-5.2 (and to a lesser extent Qwen3-max) indicate that the agreement is not merely driven by class imbalance, supporting the use of multi-judge consensus labeling for more stable and reproducible hallucination evaluation.
\begin{figure}[h]
\centering
\includegraphics[width=0.84\linewidth]{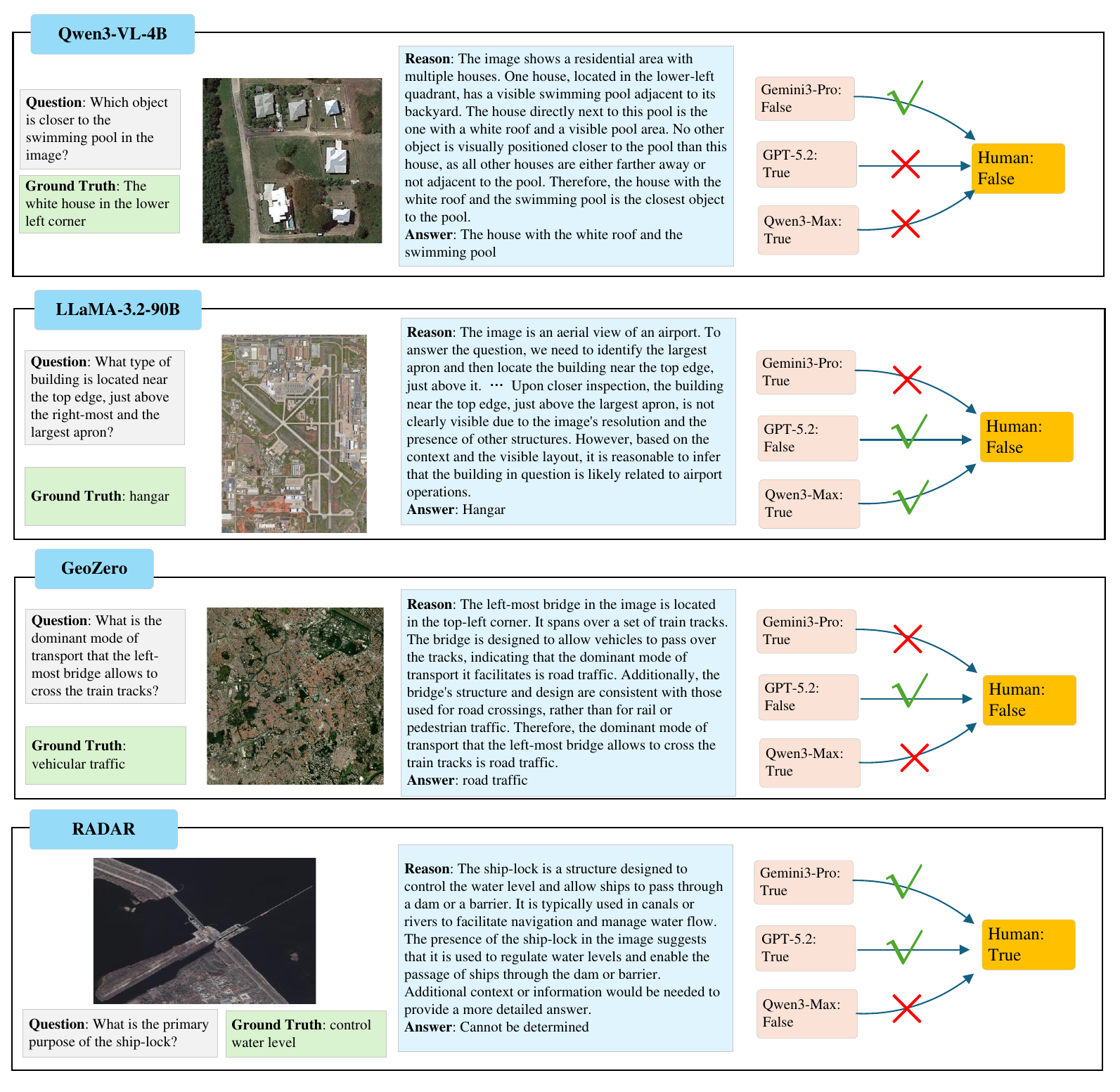}
\caption{Example from human calibration for hallucination labeling.}
\label{fig:human_clean1}
\end{figure}

\clearpage
\begin{figure}[h]
\centering
\includegraphics[width=0.84\linewidth]{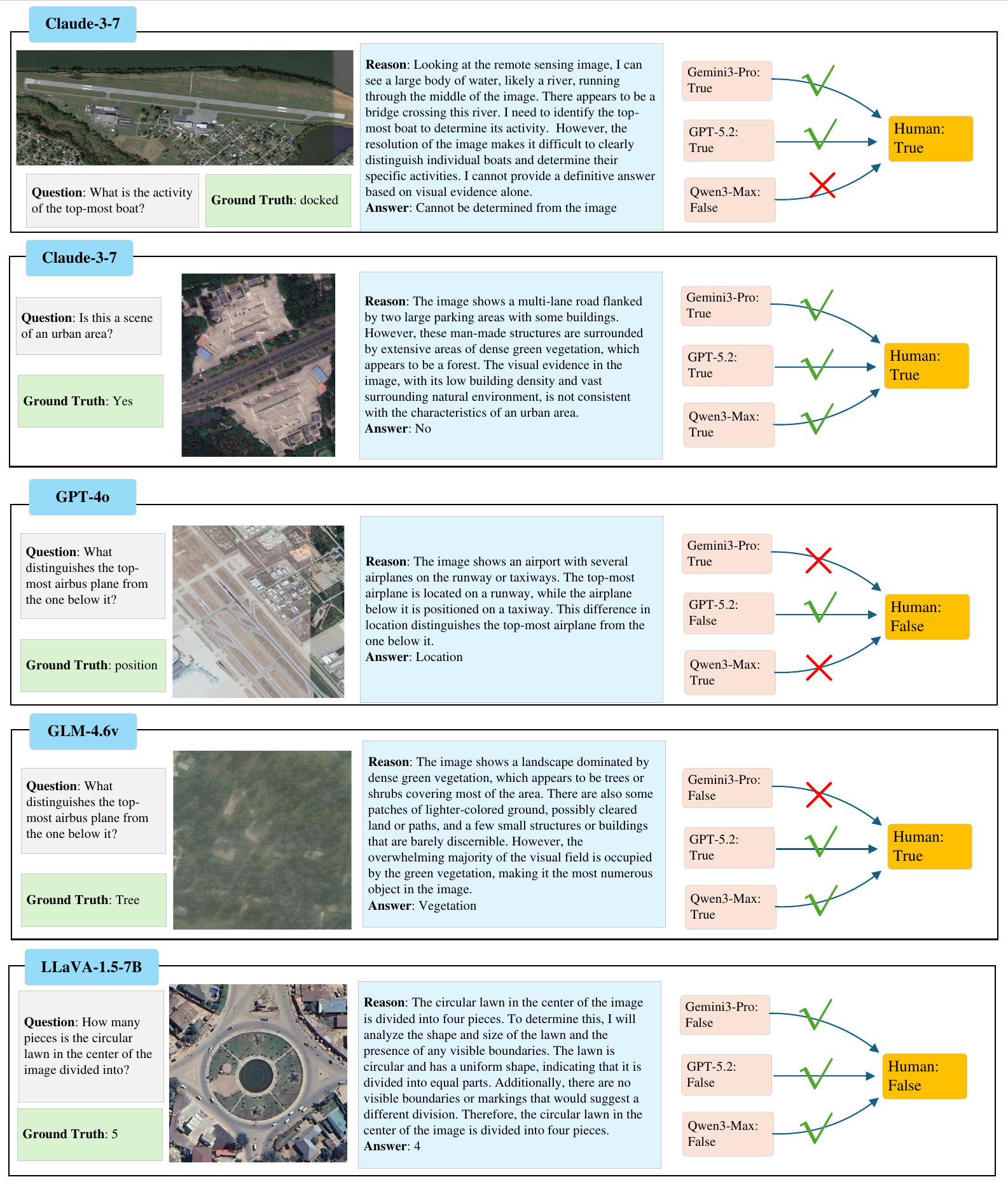}
\caption{Example from human calibration for hallucination labeling.}
\vspace{-0.4cm}
\label{fig:human_clean2}
\end{figure}

\subsubsection{Per-judge hallucination statistics}
We report per-judge hallucination statistics for each evaluated base model.
All rates are computed over 371 instances per base model.
Subtype rates are multi-label and therefore do not sum to the overall hallucination rate.

Across all three judges, hallucinations are dominated by \emph{factual} errors, where \textbf{object/category } and \textbf{attribute } hallucinations account for the largest portions across models.
In contrast, \textbf{unjustified causality} and \textbf{internal inconsistency} are consistently rare, suggesting that most failures stem from unsupported visual claims rather than explicit self-contradiction.
We also observe systematic differences in judge sensitivity.
Compared to Gemini-3-pro and Qwen3-max, GPT-5.2 assigns noticeably higher rates to \textbf{logical hallucinations}, especially \textbf{invalid reasoning} and \textbf{semantic over-attribution}, indicating stricter scrutiny on inference validity beyond surface-level factual errors.
Finally, \textbf{RADAR} yields consistent reductions under GPT-5.2 and Qwen3-max (e.g., +RADAR reduces $H$ by \(\approx 11.9\%\) and \(\approx 12.7\%\), respectively), with the largest gains coming from lower \textbf{ATT} and \textbf{SO} rates.
\clearpage
\begin{table}[h]
\centering
\caption{\textbf{Per-judge hallucination taxonomy statistics across three judges.}
All values are percentages. Subtype rates are multi-label.}
\label{tab:perjudge_merged}
\setlength{\tabcolsep}{4pt}
\renewcommand{\arraystretch}{1.08}

\resizebox{0.78\linewidth}{!}{%
\begin{tabular}{l c c c c c c c c c c}
\toprule

\multicolumn{11}{l}{\textbf{Judge: Gemini-3-pro}} \\
\midrule
\multirow{2}{*}{Models} & \multirow{2}{*}{$HR$} &
\multicolumn{4}{c}{Factual Hallucination} &
\multicolumn{5}{c}{Logical Hallucination} \\
\cmidrule(lr){3-6}\cmidrule(lr){7-11}
& & OBJ & ATT & SPA & $HR_{F}$ & IR & CI & INC & SO & $HR_{L}$ \\

\midrule
\multicolumn{11}{l}{\textit{\textcolor{gray}{Closed-source Models}}} \\
Claude-3-7      & 56.33 & 55.53 &  9.43 & 34.50 & 23.45 &  7.82 &  7.28 & 0.27 & 0.81 & 0.27 \\
Gemini-2.5-pro  & 45.28 & 44.20 &  9.70 & 18.06 & 23.18 & 11.86 &  7.55 & 0.00 & 0.54 & 0.27 \\
GPT-4o          & 45.55 & 43.40 &  7.82 & 16.98 & 22.37 &  8.89 &  5.66 & 0.27 & 0.54 & 1.08 \\

\midrule
\multicolumn{11}{l}{\textit{\textcolor{gray}{Open-source Models}}} \\
GLM-4.6v        & 47.71 & 45.28 &  9.97 & 28.03 & 17.25 &  6.74 &  8.89 & 0.00 & 0.00 & 0.00 \\
LLaVA-1.5-7B    & 66.58 & 63.61 & 19.95 & 28.03 & 27.49 &  9.97 & 15.63 & 0.81 & 2.70 & 1.89 \\
Qwen3-VL-4B     & 56.87 & 54.99 & 12.94 & 29.38 & 21.83 & 12.40 &  9.43 & 0.27 & 2.16 & 1.08 \\
LLaMA-3.2-90B   & 52.29 & 50.67 & 13.75 & 26.42 & 22.37 & 10.78 &  9.43 & 0.27 & 2.16 & 0.54 \\
GeoZero         & 64.15 & 60.92 & 14.82 & 31.81 & 28.57 & 14.29 & 10.24 & 1.08 & 1.62 & 1.62 \\
\midrule\rowcolor{gray!15}
GeoZero+RADAR   & 64.69 & 63.61 &  9.16 & 38.27 & 29.92 & 12.40 &  5.66 & 0.27 & 1.62 & 1.08 \\

\midrule\midrule

\multicolumn{11}{l}{\textbf{Judge: GPT-5.2}} \\
\midrule
\multirow{2}{*}{Models} & \multirow{2}{*}{$HR$} &
\multicolumn{4}{c}{Factual Hallucination} &
\multicolumn{5}{c}{Logical Hallucination} \\
\cmidrule(lr){3-6}\cmidrule(lr){7-11}
& & OBJ & ATT & SPA & $HR_{F}$ & IR & CI & INC & SO & $HR_{L}$ \\

\midrule
\multicolumn{11}{l}{\textit{\textcolor{gray}{Closed-source Models}}} \\
Claude-3-7      & 62.26 & 56.60 & 19.68 & 31.54 & 20.49 &  7.55 & 17.79 & 0.00 & 0.54 & 8.09 \\
Gemini-2.5-pro  & 55.80 & 52.56 & 26.68 & 27.22 & 23.45 & 10.24 & 26.15 & 0.54 & 0.54 & 8.36 \\
GPT-4o          & 50.67 & 45.55 & 18.87 & 21.83 & 19.41 &  6.47 & 17.79 & 0.00 & 0.00 & 5.93 \\

\midrule
\multicolumn{11}{l}{\textit{\textcolor{gray}{Open-source Models}}} \\
GLM-4.6v        & 53.37 & 47.98 & 20.75 & 29.38 & 14.02 &  5.39 & 19.68 & 0.00 & 0.54 & 6.47 \\
LLaVA-1.5-7B    & 48.52 & 46.09 & 23.45 & 17.52 & 19.68 & 11.32 & 21.29 & 2.70 & 1.08 & 12.13 \\
Qwen3-VL-4B     & 64.15 & 59.57 & 28.30 & 35.58 & 18.87 &  8.89 & 25.88 & 0.00 & 0.81 & 9.97 \\
LLaMA-3.2-90B   & 54.45 & 50.67 & 25.88 & 26.42 & 17.52 &  6.47 & 23.99 & 0.00 & 3.23 & 8.89 \\
GeoZero         & 52.56 & 47.98 & 28.03 & 19.41 & 24.80 &  8.36 & 25.61 & 2.70 & 1.08 & 12.13 \\
\midrule\rowcolor{gray!15}
GeoZero+RADAR   & 40.70 & 39.35 & 22.10 & 18.06 & 17.52 &  7.82 & 20.22 & 1.89 & 0.54 & 11.86 \\

\midrule\midrule

\multicolumn{11}{l}{\textbf{Judge: Qwen3-max}} \\
\midrule
\multirow{2}{*}{Models} & \multirow{2}{*}{$HR$} &
\multicolumn{4}{c}{Factual Hallucination} &
\multicolumn{5}{c}{Logical Hallucination} \\
\cmidrule(lr){3-6}\cmidrule(lr){7-11}
& & OBJ & ATT & SPA & $HR_{F}$ & IR & CI & INC & SO & $HR_{L}$ \\

\midrule
\multicolumn{11}{l}{\textit{\textcolor{gray}{Closed-source Models}}} \\
Claude-3-7      & 51.75 & 50.67 & 16.71 & 32.88 & 22.91 &  4.31 &  8.63 & 0.00 & 0.00 & 14.02 \\
Gemini-2.5-pro  & 49.60 & 47.98 & 17.52 & 28.03 & 23.45 &  5.66 & 10.51 & 0.00 & 0.00 & 13.75 \\
GPT-4o          & 49.33 & 48.52 & 12.13 & 28.57 & 19.95 &  5.39 &  6.20 & 0.00 & 0.00 & 11.05 \\

\midrule
\multicolumn{11}{l}{\textit{\textcolor{gray}{Open-source Models}}} \\
GLM-4.6v        & 46.36 & 44.47 & 10.24 & 29.11 & 14.29 &  4.04 &  5.39 & 0.00 & 0.00 &  7.82 \\
LLaVA-1.5-7B    & 48.79 & 46.36 & 16.71 & 22.10 & 18.87 &  9.97 & 11.59 & 0.27 & 0.00 & 13.75 \\
Qwen3-VL-4B     & 57.95 & 55.80 & 16.17 & 33.69 & 22.37 &  5.66 &  7.28 & 0.00 & 0.00 & 13.21 \\
LLaMA-3.2-90B   & 53.91 & 51.21 & 19.14 & 30.73 & 18.06 &  6.47 & 12.67 & 0.00 & 0.00 & 14.02 \\
GeoZero         & 47.44 & 45.01 & 19.95 & 25.34 & 23.99 &  4.85 & 14.29 & 0.54 & 0.54 & 16.44 \\
\midrule
\rowcolor{gray!15}
GeoZero+RADAR   & 34.77 & 33.42 & 13.21 & 19.41 & 16.98 &  3.23 &  9.70 & 0.27 & 0.27 & 12.13 \\

\bottomrule
\end{tabular}%
}
\end{table}

\clearpage

\subsection{ROI-Prioritized Hallucination Evaluation}

To further examine whether hallucination behavior is influenced by irrelevant global context in large-scale remote sensing imagery, we conduct an additional evaluation under an \textbf{ROI-prioritized} setting. For samples that require localized evidence to answer the question, we manually provide annotated Regions of Interest (ROI). In this setting, when an ROI is available, we supply \emph{both} (i) the cropped ROI region and (ii) the original image with the ROI bounding box highlighted, allowing expert judge models to access localized evidence while retaining global contextual information. For samples where cropping is unnecessary or no ROI is provided, we use the original image only.Overall, the evaluation follows the principle: \emph{when an ROI is available, localized evidence is explicitly emphasized through ROI crops and bounding-box annotations; otherwise, the full image is used without modification.}

Concretely, for each instance, we first query the ROI annotation (if any). If a valid ROI exists, we (i) crop the ROI from the original image and (ii) generate a version of the full image with the ROI bounding box overlaid, and provide both as visual inputs to the expert judge models. If no valid ROI exists, we directly use the full image. All other evaluation protocols remain unchanged from the main paper: each model prediction is annotated by three expert judges; the overall hallucination rate ($HR$) is determined by majority vote (two or more judges marking an instance as hallucinated); and hallucination categories/subtypes are aggregated by taking the union of tags from judges who labeled the instance as hallucinated. All values are reported as percentages over the full evaluation set (371 instances).

\paragraph{Consensus Results under ROI-Prioritized Setting.}
Table~\ref{tab:hallu_consensus_roi} reports the consensus hallucination statistics when evaluation prioritizes localized evidence.
Overall, hallucinations remain prevalent even when the visual input is restricted to the ROI: across most baselines, the overall $HR$ stays in a similar range to the full-image setting, indicating that hallucination is not solely caused by irrelevant global context.
Consistent with the main paper, factual hallucination dominates logical hallucination across both closed-source and open-source models, and the dominant factual subtypes are \texttt{OBJ} and \texttt{ATT}, suggesting persistent difficulties in object identification and attribute grounding even under localized visual evidence.

Importantly, \textbf{GeoZero+RADAR} continues to achieve the lowest hallucination rate in this setting.
Compared to GeoZero, \textbf{GeoZero+RADAR} reduces the overall $HR$ from 48.25\% to 36.93\% (an absolute reduction of 11.32 points), and reduces factual hallucination $HR_F$ from 47.98\% to 36.66\%.
This improvement is accompanied by consistent decreases in core factual subtypes (\texttt{OBJ}, \texttt{ATT}, and \texttt{SPA}), suggesting that RADAR improves grounding beyond merely benefiting from global contextual cues.

\begin{table}[h]
\small
\centering
\caption{\textbf{Hallucination evaluation on RSHBench (ROI-prioritized setting).}
Each model prediction is annotated by three expert judges with a binary hallucination rate ($HR$) and a multi-label taxonomy comprising factual subtypes (\texttt{OBJ}, \texttt{ATT}, \texttt{SPA}) and logical subtypes (\texttt{IR}, \texttt{CI}, \texttt{INC}, \texttt{SO}), along with coarse categories $HR_{F}$ (Factual) and $HR_{L}$ (Logical).
The overall $HR$ is determined by majority vote, while category and subtype labels are aggregated by taking the union of tags from judges who marked the instance as {hallucinated}.
All values are reported as percentages over the full evaluation set.}
\label{tab:hallu_consensus_roi}

\resizebox{\linewidth}{!}{%
\begin{tabular}{
l
c c c >{\columncolor{blue!15}}c
c c c c >{\columncolor{blue!15}}c
>{\columncolor{blue!15}}c
}
\toprule
\multirow{2}{*}{Models} &
\multicolumn{4}{c}{Factual Hallucination} &
\multicolumn{5}{c}{Logical Hallucination} &
\multicolumn{1}{>{\columncolor{blue!15}}c}{\textbf{HR}} \\
\cmidrule(lr){2-5}\cmidrule(lr){6-10}
& OBJ & ATT & SPA & $HR_F$ & IR & CI & INC & SO & $HR_L$
& \multicolumn{1}{>{\columncolor{blue!15}}c}{} \\

\midrule
\multicolumn{11}{l}{\textit{\textcolor{gray}{Closed-source Models}}} \\
Claude-3-7      & 49.33 & 32.61 & 16.98 & 60.11 & 19.95 & 0.00 & 0.81 & 15.09 & 24.80 & 60.11 \\
Gemini-2.5-pro  & 39.08 & 36.39 & 19.14 & 52.83 & 29.92 & 0.54 & 0.54 & 15.63 & 33.96 & 53.64 \\
GPT-4o          & 34.23 & 26.68 & 13.21 & 47.98 & 19.95 & 0.00 & 0.54 & 12.67 & 22.37 & 48.79 \\
\midrule
\multicolumn{11}{l}{\textit{\textcolor{gray}{Open-source Models}}} \\
GLM-4.6v        & 36.93 & 21.83 &  9.70 & 47.71 & 21.56 & 0.00 & 0.27 &  9.70 & 24.80 & 49.33 \\
LLaVA-1.5-7B    & 28.57 & 29.38 & 14.29 & 47.98 & 22.10 & 1.62 & 3.23 & 15.63 & 24.80 & 47.98 \\
Qwen3-VL-4B     & 49.60 & 32.88 & 17.25 & 61.73 & 27.49 & 0.27 & 1.89 & 18.87 & 33.69 & 61.99 \\
LLaMA-3.2-90B   & 43.40 & 29.11 & 14.82 & 55.53 & 32.88 & 0.00 & 4.31 & 17.25 & 36.66 & 55.80 \\
GeoZero         & 33.15 & 32.88 & 17.79 & 47.98 & 28.30 & 3.77 & 2.43 & 17.79 & 30.73 & 48.25 \\
\midrule
GeoZero\textbf{+RADAR} &
25.88 & 24.53 & 12.13 & \textbf{36.66} &
21.02 & 1.89 & 1.35 & 14.82 & 23.45 & \textbf{36.93} \\
\bottomrule
\end{tabular}%
}
\end{table}

\paragraph{Per-Judge Analysis under ROI-Prioritized Setting.}
Tables~\ref{tab:perjudge_gemini_roi}, \ref{tab:perjudge_gpt52_roi}, and \ref{tab:perjudge_qwen_roi} summarizes the per-judge hallucination statistics in a unified view.
While absolute hallucination rates vary across judges (reflecting different judging strictness), the relative trends are consistent:
\textbf{GeoZero+RADAR} yields lower $HR$ than GeoZero for all three judges (65.77 vs.\ 68.19 for Gemini-3-pro-preview; 38.81 vs.\ 49.60 for GPT-5.2; 35.85 vs.\ 46.63 for Qwen3-max), indicating that the observed gain is not driven by a single judge model.
Moreover, reductions are primarily manifested in factual hallucination components (notably \texttt{OBJ} and \texttt{ATT}), aligning with the consensus-level finding that ROI prioritization does not remove hallucinations, but RADAR consistently improves evidence faithfulness under localized visual input.

\paragraph{Comparison with Full-Image Evaluation.}
Comparing Tables~\ref{tab:hallu_consensus_roi} and Tables~\ref{tab:perjudge_gemini_roi}, \ref{tab:perjudge_gpt52_roi}, and \ref{tab:perjudge_qwen_roi} with their full-image counterparts in the main paper, we observe that restricting the visual input to ROIs does not eliminate hallucinations, and the overall model ranking remains largely stable.
In particular, the advantage of \textbf{GeoZero+RADAR} persists under ROI prioritization, suggesting that RADAR improves robustness to hallucination beyond simply mitigating background interference, and instead enhances the alignment between model outputs and localized visual evidence.

\begin{table}[h]
\centering
\caption{\textbf{Per-judge hallucination taxonomy statistics on RSHBench (ROI-prioritized setting). Judge: Gemini-3-pro-preview.} All values are percentages. Subtype rates are multi-label.}
\label{tab:perjudge_gemini_roi}
\setlength{\tabcolsep}{4pt}
\renewcommand{\arraystretch}{1.08}

\resizebox{0.78\linewidth}{!}{%
\begin{tabular}{l c c c c c c c c c c}
\toprule
\multirow{2}{*}{Models} & \multirow{2}{*}{$HR$} &
\multicolumn{4}{c}{Factual Hallucination} &
\multicolumn{5}{c}{Logical Hallucination} \\
\cmidrule(lr){3-6}\cmidrule(lr){7-11}
& & OBJ & ATT & SPA & $HR_{F}$ & IR & CI & INC & SO & $HR_{L}$ \\
\midrule
\multicolumn{11}{l}{\textit{\textcolor{gray}{Closed-source Models}}} \\
Claude-3-7      & 64.69 & 48.52 & 23.99 & 14.56 & 64.69 & 5.93 & 0.00 & 1.08 & 1.08 & 9.16 \\
Gemini-2.5-pro  & 61.46 & 35.85 & 26.42 & 20.75 & 61.19 & 9.97 & 0.27 & 0.00 & 0.54 & 12.40 \\
GPT-4o          & 54.72 & 31.00 & 20.75 & 12.40 & 53.64 & 6.47 & 0.00 & 0.54 & 0.54 & 8.63 \\
\midrule
\multicolumn{11}{l}{\textit{\textcolor{gray}{Open-source Models}}} \\
GLM-4.6v        & 56.33 & 34.23 & 18.06 & 7.55 & 54.72 & 9.70 & 0.00 & 0.27 & 0.27 & 11.86 \\
LLaVA-1.5-7B    & 64.96 & 29.92 & 26.68 & 10.78 & 61.99 & 14.56 & 1.08 & 3.23 & 3.50 & 19.95 \\
Qwen3-VL-4B     & 66.58 & 40.16 & 23.45 & 14.56 & 65.23 & 10.24 & 0.27 & 1.62 & 0.54 & 13.21 \\
LLaMA-3.2-90B   & 63.34 & 38.01 & 24.80 & 14.56 & 62.80 & 12.13 & 0.00 & 3.23 & 1.35 & 18.06 \\
GeoZero         & 68.19 & 34.77 & 28.03 & 15.63 & 66.58 & 8.09 & 1.08 & 1.62 & 1.62 & 12.94 \\
\midrule\rowcolor{gray!15}
GeoZero+RADAR   & 65.77 & 40.16 & 26.15 & 11.32 & 64.69 & 6.20 & 0.27 & 1.62 & 1.62 & 9.70 \\
\bottomrule
\end{tabular}}
\end{table}

\begin{table}[h]
\centering
\caption{\textbf{Per-judge hallucination taxonomy statistics on RSHBench (ROI-prioritized setting). Judge: GPT-5.2.} All values are percentages. Subtype rates are multi-label.}
\label{tab:perjudge_gpt52_roi}
\setlength{\tabcolsep}{4pt}
\renewcommand{\arraystretch}{1.08}

\resizebox{0.78\linewidth}{!}{%
\begin{tabular}{l c c c c c c c c c c}
\toprule
\multirow{2}{*}{Models} & \multirow{2}{*}{$HR$} &
\multicolumn{4}{c}{Factual Hallucination} &
\multicolumn{5}{c}{Logical Hallucination} \\
\cmidrule(lr){3-6}\cmidrule(lr){7-11}
& & OBJ & ATT & SPA & $HR_{F}$ & IR & CI & INC & SO & $HR_{L}$ \\
\midrule
\multicolumn{11}{l}{\textit{\textcolor{gray}{Closed-source Models}}} \\
Claude-3-7      & 61.73 & 34.50 & 22.91 & 7.01 & 58.49 & 14.82 & 0.00 & 0.27 & 7.82 & 16.98 \\
Gemini-2.5-pro  & 57.14 & 26.95 & 25.07 & 12.13 & 53.91 & 23.45 & 0.54 & 0.54 & 8.36 & 24.80 \\
GPT-4o          & 51.21 & 24.53 & 20.49 & 7.55 & 48.25 & 15.63 & 0.00 & 0.00 & 6.47 & 16.44 \\
\midrule
\multicolumn{11}{l}{\textit{\textcolor{gray}{Open-source Models}}} \\
GLM-4.6v        & 50.94 & 28.84 & 13.75 & 6.47 & 46.90 & 19.68 & 0.00 & 0.00 & 6.20 & 20.75 \\
LLaVA-1.5-7B    & 47.44 & 17.52 & 21.83 & 11.05 & 46.90 & 18.06 & 1.35 & 1.08 & 9.70 & 19.68 \\
Qwen3-VL-4B     & 64.15 & 37.74 & 21.83 & 8.36 & 61.73 & 21.02 & 0.00 & 0.27 & 11.05 & 25.07 \\
LLaMA-3.2-90B   & 56.60 & 27.22 & 20.22 & 7.28 & 52.83 & 27.22 & 0.00 & 2.96 & 8.36 & 29.11 \\
GeoZero         & 49.60 & 17.79 & 22.91 & 9.16 & 45.55 & 23.72 & 2.43 & 1.08 & 11.05 & 25.07 \\
\midrule\rowcolor{gray!15}
GeoZero+RADAR   & 38.81 & 15.36 & 18.87 & 7.01 & 37.20 & 18.60 & 1.62 & 0.81 & 9.16 & 20.49 \\
\bottomrule
\end{tabular}}
\end{table}

\begin{table}[h]
\centering
\caption{\textbf{Per-judge hallucination taxonomy statistics on RSHBench (ROI-prioritized setting). Judge: Qwen3-max.} All values are percentages. Subtype rates are multi-label.}
\label{tab:perjudge_qwen_roi}
\setlength{\tabcolsep}{4pt}
\renewcommand{\arraystretch}{1.08}

\resizebox{0.78\linewidth}{!}{%
\begin{tabular}{l c c c c c c c c c c}
\toprule
\multirow{2}{*}{Models} & \multirow{2}{*}{$HR$} &
\multicolumn{4}{c}{Factual Hallucination} &
\multicolumn{5}{c}{Logical Hallucination} \\
\cmidrule(lr){3-6}\cmidrule(lr){7-11}
& & OBJ & ATT & SPA & $HR_{F}$ & IR & CI & INC & SO & $HR_{L}$ \\
\midrule
\multicolumn{11}{l}{\textit{\textcolor{gray}{Closed-source Models}}} \\
Claude-3-7      & 52.29 & 32.08 & 21.29 & 6.20 & 50.67 & 10.24 & 0.00 & 0.00 & 15.09 & 17.52 \\
Gemini-2.5-pro  & 50.13 & 28.03 & 22.64 & 6.20 & 47.71 & 11.86 & 0.00 & 0.00 & 12.94 & 19.14 \\
GPT-4o          & 50.40 & 28.84 & 18.60 & 5.66 & 48.25 & 7.55 & 0.00 & 0.00 & 12.40 & 15.90 \\
\midrule
\multicolumn{11}{l}{\textit{\textcolor{gray}{Open-source Models}}} \\
GLM-4.6v        & 46.36 & 29.11 & 12.94 & 4.85 & 43.67 & 4.85 & 0.00 & 0.00 & 8.89 & 11.59 \\
LLaVA-1.5-7B    & 49.60 & 23.72 & 17.25 & 9.97 & 47.17 & 12.94 & 0.27 & 0.00 & 14.56 & 18.60 \\
Qwen3-VL-4B     & 56.60 & 35.04 & 18.33 & 6.74 & 53.91 & 10.24 & 0.00 & 0.00 & 14.82 & 19.14 \\
LLaMA-3.2-90B   & 54.72 & 33.96 & 15.63 & 6.20 & 51.75 & 14.29 & 0.00 & 0.00 & 16.17 & 22.10 \\
GeoZero         & 46.63 & 24.80 & 22.64 & 4.31 & 43.94 & 13.75 & 0.27 & 0.00 & 16.44 & 20.75 \\
\midrule\rowcolor{gray!15}
GeoZero+RADAR   & 35.85 & 20.49 & 15.09 & 4.85 & 35.04 & 11.32 & 0.00 & 0.00 & 13.75 & 15.63 \\
\bottomrule
\end{tabular}}
\end{table}

\section{RADAR: Additional Method and Implementation Details}
\label{app:radar_details}

\subsection{Query-Conditioned Relative Attention Extraction}
Given an image $I$ and a textual query, RADAR derives an attention-based relevance map from a multimodal large language model as an intrinsic cue for visual grounding. Let the model produce attention matrices across multiple layers. For each layer $\ell$, we summarize cross-modal attention into a spatial relevance map $A_{\ell}(Q; I)$ that reflects how strongly the query attends to each visual token, for example by aggregating over attention heads and query tokens. The resulting token-level scores are reshaped onto the image grid to obtain a layer-wise heatmap.

Absolute attention is often dominated by scene-level saliency that is weakly related to the query semantics. To suppress such query-irrelevant signals, we contrast the task query $Q^{T}$ with a global scene query $Q^{G}$. At each layer, we compute a stabilized relative attention map
\begin{equation}
\hat{A}_{\ell}(Q^{T}; I)=\frac{A_{\ell}(Q^{T}; I)}{A_{\ell}(Q^{G}; I)+\epsilon},
\end{equation}
where $\epsilon$ is a small constant for numerical stability. This formulation emphasizes regions whose relevance is selectively amplified by the task query relative to holistic scene understanding.

We then identify layers where this contrast is most informative. Each layer is scored by its total attention mass
\begin{equation}
s_{\ell}=\sum_{u}\hat{A}_{\ell}(u),
\end{equation}
after which the top-$k$ layers $\mathcal{S}_{k}$ are selected. For each selected layer, a normalized weight is computed as
\begin{equation}
w_{\ell}=\frac{s_{\ell}}{\sum_{j\in\mathcal{S}_{k}} s_{j}}, \quad \ell\in\mathcal{S}_{k}.
\end{equation}
The final query-conditioned relative attention map is obtained through weighted aggregation
\begin{equation}
\tilde{A}(Q^{T}; I)=\mathcal{N}\!\left(\sum_{\ell\in\mathcal{S}_{k}} w_{\ell}\,\hat{A}_{\ell}(Q^{T}; I)\right),
\end{equation}
where $\mathcal{N}(\cdot)$ normalizes the result into a spatial probability distribution.

\subsection{Focus Test and Region Mapping}
\subsubsection{Focus test}
RADAR avoids cropping when attention is diffuse.
We measure concentration using a lightweight focus test $\mathcal{F}(\cdot)$ defined on the normalized map $\tilde{A}$.
A simple instantiation uses normalized entropy:
\begin{equation}
H(\tilde{A})=-\sum_{u}\tilde{A}(u)\log\left(\tilde{A}(u)+\delta\right),
\end{equation}
where $\delta$ avoids numerical issues.
Let $N$ be the number of spatial cells.
We define a concentration score:
\begin{equation}
\mathcal{F}(\tilde{A})=1-\frac{H(\tilde{A})}{\log N}.
\end{equation}
RADAR proceeds with cropping if $\mathcal{F}(\tilde{A})\ge \tau$; otherwise, it falls back to full-image inference.

\subsubsection{Region mapping operator}
Given a focused attention map, we extract a crop using a deterministic operator $\Psi(\cdot)$.
We select the smallest set of cells whose cumulative probability mass reaches a target ratio $p$.
We then compute the tight bounding box covering these cells on the grid, map it back to pixel coordinates, and optionally add a small padding margin.
This yields the crop used for progressive evidence acquisition in Stage1 and Stage2.

\subsection{Where and What Prompt Templates}
\label{app:OPT}
We employ a protocol that converts an input question into a \emph{where} question for coarse localization and a \emph{what} question for fine-grained inspection.

\begin{promptbox}{Where and What Protocol}
\begin{PromptVerbatim}
You are a helpful assistant that decomposes an image-related question
into two simple questions to support visual understanding.
The two questions serve different purposes:
- The first question helps the model know WHERE to look in the image.
- The second question helps the model understand WHAT details to examine in that area.
Given an input question about an image, generate exactly TWO questions:
1. WHERE QUESTION (coarse visual attention)
Goal:
- Simplify the original question as much as possible.
- Help the model quickly identify the relevant region in the image.
Guidelines:
- Focus on the main target object(s) or land-cover type(s).
- Keep the question short and concrete.
- Include only information necessary to locate the region.
- You MAY use common remote-sensing associations if they are obvious and helpful.
- Do NOT ask about quantity, attributes, or detailed reasoning.
- Do NOT invent image-level positions.
2. WHAT QUESTION (fine-grained understanding)
Goal:
- Ask the original question again, assuming the correct region is known.
Guidelines:
- Preserve the original intent and question type.
- Remove only global image-position words if they exist.
- Keep the question natural and easy to understand.
- Do NOT add new information.
Output format (JSON ONLY):
{
  "where_question": "...",
  "what_question": "..."
}
Now process the following question:
\end{PromptVerbatim}
\end{promptbox}

\begin{figure*}[h]
\centering
\includegraphics[width=0.88\linewidth]{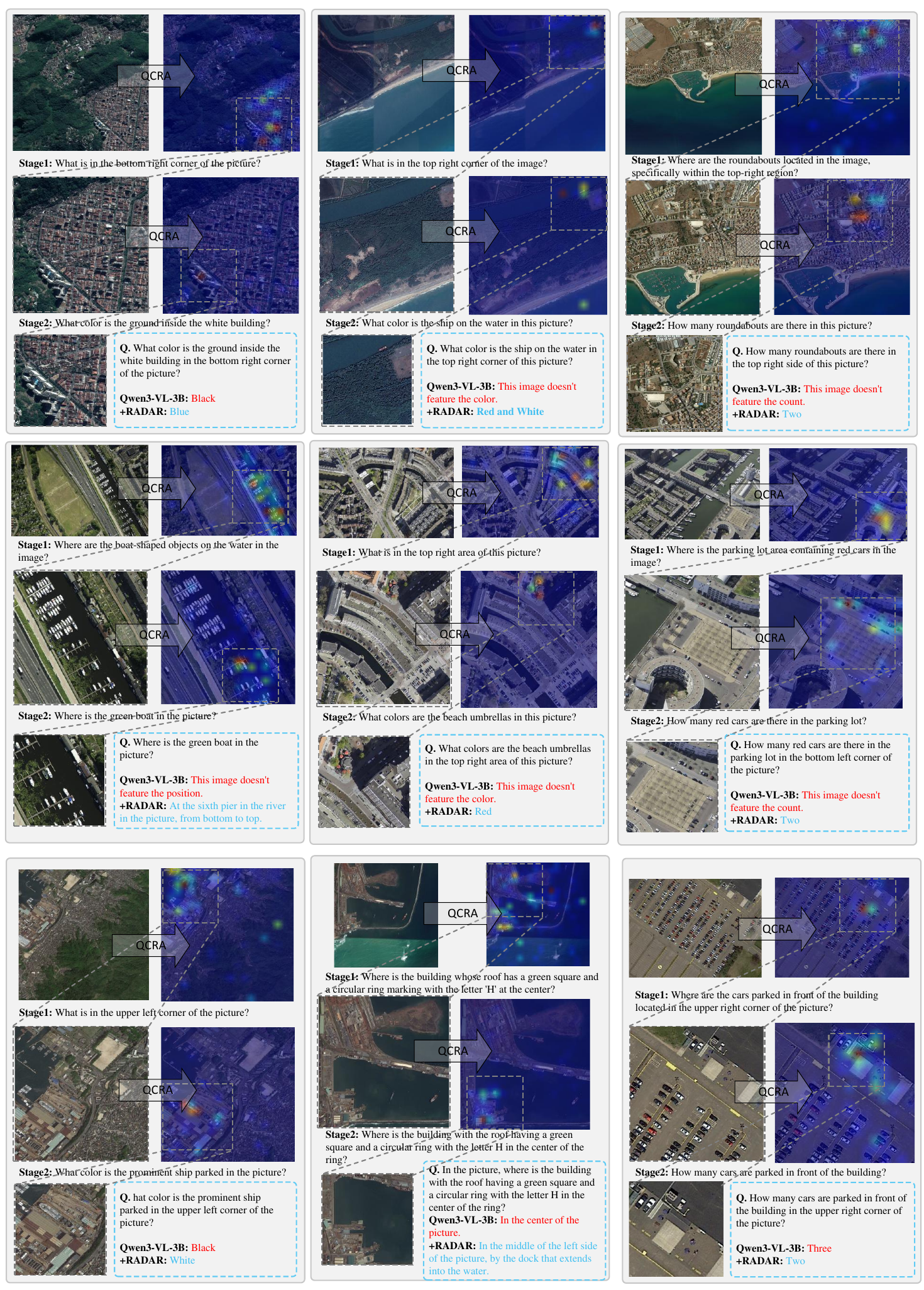}
\caption{\textbf{Qualitative examples of RADAR's query-conditioned relative attention and progressive evidence refinement.}
For each case, we visualize the QCRA heatmap produced by a \emph{where}-oriented query on the full image (Stage1) and a \emph{what}-oriented query on the localized crop (Stage2), where brighter regions indicate stronger question-conditioned relevance.
Dashed boxes denote the regions selected for zoom-in evidence extraction.}
\label{fig:qualitative11}
\end{figure*}

\end{document}